\PassOptionsToPackage{table}{xcolor}
\documentclass[acmsmall,screen,nonacm]{acmart}

\AtBeginDocument{%
  }

\usepackage{multirow} 
\usepackage{graphicx}
\usepackage{float}
\usepackage{array}
\usepackage{listings}
\usepackage{mathrsfs}

\usepackage{booktabs}
\usepackage{dsfont}
\usepackage{mathtools}
\usepackage{algorithm}
\usepackage{algpseudocode}

\usepackage[english]{babel}
\usepackage[table]{xcolor}
\usepackage{tabularx}
\usepackage{placeins}
\usepackage{amsthm}
\usepackage{xfrac}
\usepackage{nicefrac}
\usepackage{gensymb}
\usepackage{balance}
\usepackage[edges]{forest}
\usepackage{ragged2e}
\usepackage{makecell}
\usepackage{placeins}
\usepackage{tabularx}

\hypersetup{
  pdfencoding=auto,
  psdextra,
  breaklinks,
  colorlinks
}

\begin{document}

\title{Embodied Foundation Models at the Edge: A Survey of Deployment Constraints and Mitigation Strategies }

\author{Utkarsh Grover}
\affiliation{%
\institution{University of South Florida}
\city{Tampa}
\state{FL}
\postcode{33620}
\country{USA}}
\email{utkarshgrover@usf.edu}

\author{Ravi Ranjan}
\affiliation{%
\institution{Florida International University }
\city{Miami}
\state{FL}
\postcode{33199}
\country{USA}}
\email{rkuma031@fiu.edu}

\author{Mingyang Mao}
\affiliation{%
\institution{University of South Florida}
\city{Tampa}
\state{FL}
\postcode{33620}
\country{USA}}
\email{mmao@usf.edu}

\author{Trung Tien Dong}
\affiliation{%
\institution{University of South Florida}
\city{Tampa}
\state{FL}
\postcode{33620}
\country{USA}}
\email{dongt@usf.edu}

\author{Satvik Praveen}
\affiliation{%
\institution{University of South Florida}
\city{Tampa}
\state{FL}
\postcode{33620}
\country{USA}}
\email{satvikpraveen@usf.edu}

\author{Zhenqi Wu}
\affiliation{%
\institution{University of South Florida}
\city{Tampa}
\state{FL}
\postcode{33620}
\country{USA}}
\email{zhenqi@usf.edu}

\author{J. Morris Chang}
\affiliation{%
\institution{University of South Florida}
\city{Tampa}
\state{FL}
\postcode{33620}
\country{USA}}
\email{chang5@usf.edu}

\author{Tinoosh Mohsenin}
\affiliation{\institution{Johns Hopkins University}
\city{Baltimore}
\state{MD}
\postcode{21218}
\country{USA}}
\email{tinoosh@jhu.edu}

\author{Yi Sheng}
\affiliation{%
\institution{University of South Florida}
\city{Tampa}
\state{FL}
\postcode{33620}
\country{USA}}
\email{sheng1@usf.edu}

\author{Agoritsa Polyzou}
\affiliation{
\institution{Florida International University}
\city{Miami}
\state{FL}
\postcode{33199}
\country{USA}}
\email{apolyzou@fiu.edu}

\author{Eiman Kanjo}
\affiliation{%
  \institution{Nottingham Trent University}
  \city{Nottingham}
  \postcode{NG1 4FQ}
  \country{United Kingdom}}
\email{eiman.kanjo@ntu.ac.uk}
\affiliation{%
  \institution{Imperial College London}
  \city{London}
  \country{United Kingdom}}
\email{e.kanjo@Imperial.ac.uk}

\author{Xiaomin Lin}
\affiliation{%
\institution{University of South Florida}
\city{Tampa}
\state{FL}
\postcode{33620}
\country{USA}}
\email{xlin2@usf.edu}

\renewcommand{\shortauthors}{Grover et al.}

\begin{abstract}
Deploying foundation models in embodied edge systems is fundamentally a systems problem, not just a problem of model compression. Real-time control must operate within strict size, weight, and power constraints, where memory traffic, compute latency, timing variability, and safety margins interact directly. The Deployment Gauntlet organizes these constraints into eight coupled barriers that determine whether embodied foundation models can run reliably in practice. Across representative edge workloads, autoregressive Vision-Language-Action policies are constrained primarily by memory bandwidth, whereas diffusion-based controllers are limited more by compute latency and sustained execution cost. Reliable deployment therefore depends on system-level co-design across memory, scheduling, communication, and model architecture, including decompositions that separate fast control from slower semantic reasoning.

\end{abstract}

\maketitle
\section{Introduction}

The migration of Foundation Models (FMs)~\cite{qazi2025prism} from hyperscale data centers to resource-constrained edge platforms changes the execution model of intelligence itself~\cite{bommasani2021opportunities, li2023personal}. Cloud scale FMs are typically developed under assumptions of elastic power, abundant cooling, and comparatively relaxed latency constraints. Embodied platforms violate all three. Autonomous mobile robots (AMRs), aerial systems, and wearables must execute perception, inference, and control within strict Size, Weight, and Power (SWaP) budgets and under hard real-time constraints~\cite{duan2022survey}. In this regime, deploying foundation models is not merely a model compression problem, but a closed-loop systems problem~\cite{reddi2020mlperf}: memory intensive inference must remain reliable despite thermal variation, battery-limited operation, shared-memory contention, and timing-sensitive control.

This setting makes Moravec's Paradox operational rather than philosophical. High level reasoning has become increasingly accessible to large models, but robust sensorimotor integration remains computationally expensive and architecturally brittle~\cite{moravec1988mind}. Embodied autonomy depends on multimodal perception that aligns light, sound, and geometry tightly enough to preserve control fidelity~\cite{girdhar2023imagebind}. Once these multi-rate, asynchronous sensor streams are coupled to large, memory-heavy foundation models, the dominant bottlenecks shift from model accuracy alone to timing, bandwidth, synchronization, and runtime stability. These bottlenecks are far less severe in disembodied or cloud-centric deployments.

Accordingly, this survey examines embodied foundation models through the systems constraints that govern whether they can be deployed reliably on edge platforms. We do not survey the edge stack in full, nor do we treat middleware, embedded operating systems, or sensor hardware as independent topics. We consider them only where they become first-order constraints on closed-loop FM deployment. Our central claim is that many of the most consequential embodied deployment failures arise not from parameter count alone, but from cross-layer interactions among sensing, compute, memory, timing, power, and control.

\subsection{The Multimodal Execution Imperative}

Closed-loop autonomy requires world representations that can be updated and acted upon within tight latency budgets. Vision-Language-Action architectures such as RT-2 and OpenVLA pursue this goal by combining perception, semantic grounding, and policy generation in a unified inference stack~\cite{zitkovich2023rt,kim2024openvla}. That unification simplifies interfaces and improves cross-modal reasoning, but it also places sensing, inference, and control on the same critical path. A representative 7B-parameter VLA may require sustained high-bandwidth weight movement, dense attention computation, and tightly synchronized ingestion of heterogeneous sensor streams~\cite{driess2023palm}. These requirements often exceed the latency, bandwidth, and scheduling margins available on embedded edge platforms.

The mismatch is modality specific. On accelerators such as the NVIDIA Jetson AGX Orin and Qualcomm RB5, sparse LiDAR workloads underutilize dense tensor hardware and often incur CPU-mediated scatter-gather overheads~\cite{choy20194d,tang2020searching}. Transformer based audio models can exceed the SRAM capacity of low power DSPs, forcing costly off chip accesses~\cite{radford2023robust,zhang2018deep}. High resolution visual encoders generate bursty traffic that can saturate LPDDR channels and trigger thermal throttling~\cite{dosovitskiy2020image,deiana2022}. More broadly, design trends that favor large, modality spanning foundation models in server class settings~\cite{reed2022generalist,team2023gemini} become substantially harder to sustain at the sensor edge, where parameter scale, deep attention stacks, and autoregressive decoding amplify memory pressure, latency variance, and scheduling instability~\cite{sze2017efficient}. These execution pressures motivate a systems view of embodied FM deployment.

\subsection{The Deployment Gauntlet}

Most current approaches to deploying foundation models on constrained hardware emphasize local optimizations such as quantization and pruning~\cite{gholami2021survey,kulkarni2022survey}. These methods reduce inference cost and remain valuable, but they do not resolve the dominant failures of embodied deployment, where sensing, computation, memory, timing, and control interact under closed-loop execution. In embodied systems, stress in one layer readily propagates across the stack and degrades overall reliability. To organize this problem, we introduce the \textbf{\emph{Deployment Gauntlet}}, a systems taxonomy of the conditions that embodied foundation models must satisfy to operate robustly in real-time edge environments.

\begin{figure}[t]  
    \centering
    \includegraphics[width=0.9\textwidth]{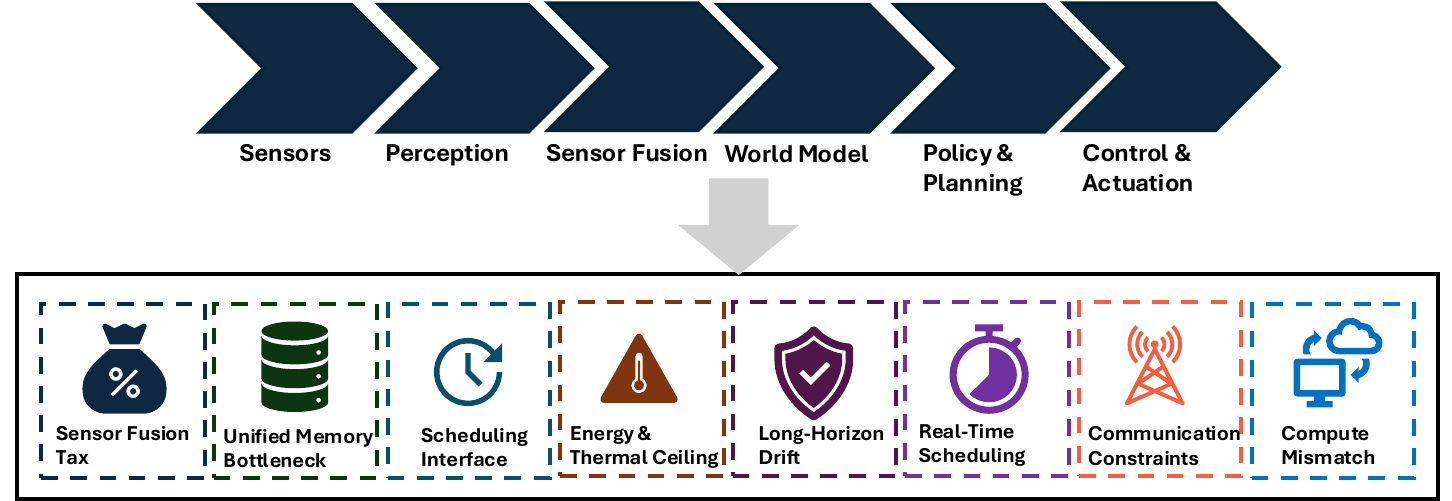} 
    \caption{The Deployment Gauntlet. A unified view of the eight major system barriers that limit the deployment of foundation models from the cloud to edge embodied AI platforms.}

    \label{fig:deployment_gauntlet}
\end{figure}

Figure~\ref{fig:deployment_gauntlet} summarizes the eight barriers that constitute the Deployment Gauntlet. They capture recurring bottlenecks across the embodied intelligence pipeline, spanning multimodal sensing and data movement, heterogeneous computation and shared memory, time critical control, long horizon operation, safety enforcement, and edge-cloud coordination. Although analytically distinct, they are tightly coupled in practice: pressure in one dimension often amplifies failure in others. The taxonomy therefore provides a concrete framework for reasoning about the deployment constraints that govern real time embodied AI.

We organize the Deployment Gauntlet into eight coupled barriers spanning the sensing to control continuum:
\begin{enumerate}
    \item \textit{The Sensor Fusion Tax:} Latency and jitter introduced by temporally aligning asynchronous sensor streams, including middleware-induced serialization and buffering overheads (e.g., ROS~2)~\cite{macenski2022robot,kronauer2021latency}.
    \item \textit{Heterogeneous Compute Mismatch:} Inefficiencies that arise when sparse or irregular workloads are mapped onto accelerators optimized for dense, static matrix operations~\cite{hooker2020hardware}.
    \item \textit{Unified Memory Bottleneck:} Contention between high-bandwidth model weight streaming and continuous sensor ingestion over shared memory channels~\cite{wang2024confidential,park2024case}.
    \item \textit{Energy \& Thermal Ceiling:} Performance degradation caused by thermal throttling and battery depletion during sustained operation~\cite{cao2020overview}.
    \item \textit{Long-Horizon Drift:} Progressive degradation of state estimation, calibration, and clock synchronization over extended deployments~\cite{cadena2016past}.
    \item \textit{Safety \& Verification Gap:} Safety-critical failures induced by hallucinated states or out-of-distribution behavior within control loops~\cite{amodei2016concrete,xiao2023motion}.
    \item \textit{Real-Time Scheduling Interference:} Control instability driven by OS jitter, non-deterministic kernel execution, and contention on shared accelerators~\cite{capodieci2022s}.
    \item \textit{Communication Constraints:} Bandwidth and latency limits that restrict edge-to-cloud offloading and multi-agent coordination~\cite{shi2016edge}.
\end{enumerate}

The remainder of the survey develops this argument in three parts. We first examine the workload characteristics of embodied foundation models and show why multimodal, closed-loop execution imposes qualitatively different demands from static cloud inference. We then use the Deployment Gauntlet to organize the principal systems barriers reported across the literature. Finally, we synthesize the mitigation strategies and architectural directions that appear most promising for improving reliability under real-time edge constraints.
\section{Landscape of Foundation Models on the Edge}
\label{sec:landscape}
\begin{table}[t]
\centering
\caption{Edge-relevant workload classes for embodied foundation model deployment.}
\label{tab:workload_taxonomy}
\scriptsize
\setlength{\tabcolsep}{3.5pt}
\renewcommand{\arraystretch}{1.12}
\begin{tabularx}{\columnwidth}{|m{1.8cm}|m{2.2cm}|m{2.0cm}|X|}
\hline
\rowcolor[HTML]{800000}
{\color{white}\textbf{Workload Class}} &
{\color{white}\textbf{Representative Models}} &
{\color{white}\textbf{Primary Role / Modality}} &
{\color{white}\textbf{Dominant Bottleneck}} \\
\hline\hline

Vision-Language-Action (VLA) Policies &
RT-2~\cite{zitkovich2023rt}, OpenVLA~\cite{kim2024openvla}, NanoVLA~\cite{chen2025nanovla} &
Policy generation; RGB + language to action &
Memory traffic and weight streaming during autoregressive decoding, which compete directly with sensor DMA and shared-memory service. \\
\hline

Diffusion-Based Policies &
Octo~\cite{team2024octo}, OneDP~\cite{wang2024one}, LightDP~\cite{wu2025device} &
Trajectory generation; RGB / multimodal to action sequence &
Sustained compute and thermal load from iterative denoising, which imposes a latency floor and high energy cost. \\
\hline

Vision Encoders and Multimodal LMMs &
LLaVA-Mini~\cite{zhang2025llava}, MiniCPM-V~\cite{yao2025efficient} &
Perception and reasoning; RGB + language &
Visual prefill and bursty frame ingestion, which create transient compute and memory spikes during high-resolution encoding. \\
\hline

3D / LiDAR Encoders &
PointPillars~\cite{lang2019pointpillars}, DensePillarNet~\cite{chandorkar2025rethinking}, BEVFusion~\cite{liu2022bevfusion} &
Geometric perception; LiDAR / LiDAR + RGB &
Sparsity and irregular memory access, which reduce accelerator utilization and increase synchronization overhead. \\
\hline

Multimodal Fusion Stacks &
MMEdge~\cite{huang2025mmedge}, BEVFusion~\cite{liu2022bevfusion} &
Cross-modal state construction; RGB / LiDAR / audio / text &
Scheduling and synchronization pressure from shared middleware, memory bandwidth, and callback timing. \\
\hline

\end{tabularx}
\end{table}

\FloatBarrier
Table~\ref{tab:workload_taxonomy} summarizes the workload classes most relevant to embodied foundation model deployment at the edge, together with their representative models, functional role, and dominant systems bottleneck.
Embodied deployment increasingly favors decomposed execution over monolithic end-to-end inference. In practice, server-resident cognitive backbones are paired with compressed on-device executors that handle latency-critical perception and control. Consumer systems already reflect this pattern. Apple’s iOS~18 ``Intelligence,'' for example, couples a server-resident model with an on-device model of roughly three billion parameters, using LoRA adapters and mixed 2-bit/4-bit quantization to preserve interactive latency~\cite{niu2025collaborative}. Google’s \emph{Gemini Robotics On-Device} likewise executes vision-language-action (VLA) policies entirely on board for bimanual manipulation without network dependence~\cite{team2025gemini}.

Robotic platforms make the same constraints more severe. Foundation level policies such as PaLM-SayCan and RT-2 demonstrate strong semantic grounding and task planning~\cite{Ahn2022DoAI,zitkovich2023rt}, but deployment still depends on distillation, hardware specialization, or architectural decomposition to satisfy latency and power limits. Recent examples include diffusion-transformer-based Large Behavior Models on Boston Dynamics’ Atlas~\cite{zeng2025behavior} and the deployable on-board inference engines used in NVIDIA’s R\textsuperscript{2}D\textsuperscript{2} sim-to-real pipeline~\cite{yu2025real2render2real}. The implication is clear: unconstrained generative inference does not by itself yield safe embodied execution. Reliable deployment requires co-design across model architecture, training strategy, runtime, and embedded hardware~\cite{niu2025collaborative}.

This section focuses on the workload classes that most clearly expose the systems pressures later organized by the Deployment Gauntlet. The objective is not to catalog every edge workload in embodied AI, but to identify the classes whose execution is dominated by distinct bottlenecks under Size, Weight, and Power (SWaP) constraints. We therefore focus on workloads governed primarily by memory bandwidth, sustained compute demand, visual prefill cost, sparsity and irregular access, or cross-modal scheduling pressure.

\subsection{Taxonomy of Edge-Relevant Workloads}

Aggregate metrics such as parameter count and FLOPs are poor predictors of whether a foundation model will execute reliably on an embodied edge platform~\cite{hooker2020hardware}. We therefore classify edge-relevant workloads along three practical axes. The first is the \textit{primary role} of the model in the control loop, distinguishing perception-heavy models from policy-generating models~\cite{Firoozi2023FoundationMI, li2025comprehensive}. The second is the \textit{I/O modality}, which ranges from dense, high-throughput inputs such as RGB video to sparse, irregular inputs such as LiDAR point clouds~\cite{xu2023multimodal}. The third is the \textit{dominant execution bottleneck}: memory capacity and bandwidth, sustained compute demand, or I/O throughput~\cite{williams2009roofline, prashanthi2025pagoda}.

These axes matter because workloads with similar nominal scale can behave very differently once deployed under SWaP and real-time constraints. Two models with comparable parameter counts may place entirely different demands on memory traffic, thermal headroom, accelerator utilization, or scheduler stability. For this reason, the discussion below emphasizes workload structure rather than aggregate model size.

\subsection{Vision-Language-Action (VLA) Policies}

Vision-Language-Action (VLA) policies~\cite{goyal2025vla, sapkota2025vision} are among the most demanding workloads in embodied AI because they couple semantic reasoning directly to action generation. Architectures such as RT-2~\cite{zitkovich2023rt} and OpenVLA~\cite{kim2024openvla} jointly tokenize observations and actions, producing discrete control outputs such as end-effector poses or skill tokens from RGB inputs and language commands. This unification improves task generalization, but it also places deliberation and actuation on the same execution path. OpenVLA-7B, for example, requires roughly 14\,GB in FP16, which approaches or exceeds the practical memory budget of unified-memory SoCs such as the Jetson Orin NX once operating system overhead and sensor buffering are accounted for.

At the edge, VLA execution is constrained primarily by memory traffic rather than peak arithmetic throughput. Autoregressive decoding repeatedly streams large weight tensors for each generated action token, creating sustained DRAM pressure. On embedded platforms, this traffic competes directly with high-rate sensor DMA streams and can introduce latency spikes that violate control deadlines~\cite{prashanthi2025pagoda}. Monolithic VLA inference is therefore difficult to reconcile with deterministic closed-loop control.

Recent systems address this mismatch by decomposing VLA inference across timescales. NanoVLA~\cite{chen2025nanovla}, for example, assigns high-level semantic reasoning to a heavyweight vision-language model while delegating higher-frequency control to a lightweight visual policy. This separation reduces sustained memory pressure and substantially improves execution speed on edge hardware. For the Deployment Gauntlet, VLA policies are best viewed as a \emph{memory-bandwidth-dominated} workload, and they directly motivate the barriers associated with shared memory contention and real-time interference~\cite{Ahn2022DoAI}.

\subsection{Diffusion-Based Policies}

Diffusion-based policies define a different edge regime. Rather than emitting discrete action tokens, they generate continuous action trajectories through iterative denoising~\cite{chi2025diffusion}. This structure is attractive for contact-rich manipulation and smooth trajectory generation, and systems such as Octo~\cite{team2024octo} illustrate its promise for generalist multimodal behavior. The same iterative structure, however, imposes a latency floor that conflicts with the update rates expected in real-time control.

On edge hardware, diffusion policies are constrained primarily by sustained compute demand and energy draw. Each denoising step performs dense matrix operations over the same control horizon, keeping accelerators active for long intervals and increasing both power and thermal load. Naive multi-step diffusion controllers often operate at only 1--2\,Hz, which is inadequate for many closed-loop manipulation settings. The main deployment question is therefore not whether diffusion can produce high-quality trajectories, but whether it can do so within a timing and thermal envelope compatible with embodied control.

Edge-oriented work has focused on collapsing the denoising process. OneDP~\cite{wang2024one} distills multi-step diffusion into a single-step generator and reports an increase from roughly 1.5\,Hz to more than 60\,Hz on representative edge hardware. LightDP~\cite{wu2025device} reduces compute demand through pruning and architectural simplification to stay within mobile thermal limits. These methods improve responsiveness substantially, but they do not change the fact that diffusion policies remain a \emph{compute and energy-dominated} workload. In the Deployment Gauntlet, they map most directly to the energy, thermal, and accelerator utilization barriers.

\subsection{Vision Encoders and Multimodal LMMs}

Vision encoders and multimodal large language models (LMMs) impose a third execution profile, dominated by the cost of ingesting high-resolution visual streams at video rate. These models provide the perceptual substrate for downstream planners and controllers, but at the edge their limiting factor is often not steady-state decoding. It is repeated visual \emph{prefill}. Large encoders such as CLIP-ViT-L provide strong semantic representations~\cite{radford2023robust}, yet their quadratic attention cost makes naive frame-rate deployment computationally and thermally expensive on embedded hardware.

Edge oriented perception pipelines therefore aim to compress or amortize visual ingestion. LLaVA-Mini~\cite{zhang2025llava} reduces prefill cost by compressing visual input to a single token before language-model ingestion. MiniCPM-V~\cite{yao2025efficient} shows that architecture choices tuned explicitly for mobile inference, combined with targeted data curation, can approach high-end multimodal performance without increasing model scale. ViT-Edge~\cite{saha2025vision} further systematizes encoder-side optimizations such as low-rank factorization and structured compression.

For the Deployment Gauntlet, the central point is that these models are \emph{prefill-bound} and thermally sensitive. Continuous video ingestion creates bursty compute demand during frame processing, and those bursts can dominate the power envelope of fanless edge modules~\cite{prashanthi2025pagoda}. When visual tokens cannot be cached, reused, or updated incrementally across frames, the encoder becomes a major source of throttling even if downstream reasoning is lightweight. This workload therefore motivates the barriers associated with bandwidth spikes, thermal instability, and burst-driven runtime interference.

\subsection{3D \& LiDAR Encoders}

3D and LiDAR encoders expose a different systems limitation: sparsity. Unlike dense vision workloads, point-cloud pipelines are dominated by irregular memory access, dynamic spatial density, and data-dependent control flow. These properties violate the locality and throughput assumptions built into most dense accelerators~\cite{choy20194d}. PointPillars remains a widely used baseline because it projects sparse point clouds into pseudo-images and thereby recovers partial compatibility with 2D CNN hardware~\cite{lang2019pointpillars}.

That projection only mitigates the problem. DensePillarNet~\cite{chandorkar2025rethinking} restructures the backbone to favor denser execution and reduces end-to-end latency on constrained devices, but voxelization and feature aggregation still introduce irregular accesses that are difficult to schedule efficiently. In multimodal systems such as BEVFusion~\cite{liu2022bevfusion}, intermediate features must often be scattered and gathered across CPU and GPU boundaries, creating synchronization overhead and latency variance.

From an execution perspective, 3D perception is \emph{sparsity-bound}. Voxelization and point aggregation disrupt cache behavior, reduce SIMD utilization, and map poorly to fixed-function inference engines such as NVIDIA’s DLA, which lack native support for sparse indexing and dynamic structures~\cite{tang2020searching}. As a result, even moderately sized 3D perception stacks can monopolize shared compute and memory resources and interfere with time-critical control. Within the Deployment Gauntlet, these workloads motivate the barriers associated with heterogeneous compute mismatch and shared-resource contention.
Table~\ref{tab:edge_workloads_a} summarizes the workload classes most relevant to embodied foundation model deployment at the edge, together with the dominant execution bottleneck and the characteristic failure mode each class introduces under SWaP and real-time constraints.

\begin{table}[t]
\centering
\caption{Representative edge-relevant workload classes and their dominant execution bottlenecks in embodied foundation model deployment.}
\label{tab:edge_workloads_a}
\scriptsize
\setlength{\tabcolsep}{4pt}
\renewcommand{\arraystretch}{1.15}
\begin{tabularx}{\columnwidth}{|p{1.95cm}|p{2.15cm}|X|}
\hline
\rowcolor[HTML]{800000}
{\color{white}\textbf{Workload}} &
{\color{white}\textbf{Dominant Bottleneck}} &
{\color{white}\textbf{Characteristic Edge Failure Mode}} \\
\hline\hline

Vision-Language-Action (VLA) Policies
& Memory traffic and weight streaming
& Autoregressive action decoding repeatedly streams large parameter tensors, competing with sensor DMA traffic over shared memory and introducing latency spikes that can violate closed-loop control deadlines~\cite{kim2024openvla,prashanthi2025pagoda,chen2025nanovla}. \\
\hline

Diffusion-Based Policies
& Sustained compute and thermal load
& Iterative denoising requires repeated dense matrix execution over the same control horizon, creating a latency floor, high energy draw, and thermal throttling that reduce practical control frequency on edge platforms~\cite{chi2025diffusion,wang2024one,wu2025device}. \\
\hline

Vision Encoders and Multimodal LMMs
& Visual prefill and bursty ingestion
& High-resolution visual streams require repeated encoder prefill, producing burst-driven compute and memory demand that can dominate the thermal envelope even when downstream reasoning remains lightweight~\cite{radford2023robust,zhang2025llava,yao2025efficient,saha2025vision}. \\
\hline

3D \& LiDAR Encoders
& Sparsity and irregular memory access
& Point-cloud pipelines rely on voxelization, sparse indexing, and data-dependent aggregation, which map poorly to dense accelerators, reduce utilization, and introduce synchronization overhead in multimodal stacks~\cite{choy20194d,lang2019pointpillars,tang2020searching,liu2022bevfusion}. \\
\hline

Multimodal Fusion Stacks
& Scheduling and synchronization
& Even when unimodal components meet latency targets in isolation, their composition through shared middleware, LPDDR bandwidth, and runtime scheduling can produce temporal misalignment and end-to-end timing failures~\cite{feng2020deep,kronauer2021latency,huang2025mmedge,bechtel2024analysis}. \\
\hline

\end{tabularx}
\end{table}
\FloatBarrier
\subsection{Multimodal Fusion Stacks}

Multimodal fusion is where these workload-specific pressures converge. Fusion stacks combine heterogeneous perceptual streams into a temporally consistent state for planning and control~\cite{feng2020deep}. They do not introduce a single new compute primitive so much as couple the execution behavior of all upstream modalities. Fusion is therefore less a model-level bottleneck than a systems-level one.

Recent systems such as MMEdge~\cite{huang2025mmedge} replace sequential fusion with pipelined sensing that overlaps ingestion, encoding, and fusion. This improves average throughput, but throughput alone is insufficient. Once modalities share middleware, memory bandwidth, and runtime scheduling, end-to-end behavior depends on callback latency, executor design, DMA contention, and operating-system preemption~\cite{kronauer2021latency}. On unified-memory SoCs, fusion also becomes the locus of contention among sensor ingestion, model weight movement, and intermediate activation storage~\cite{bechtel2024analysis}. 
Parallel multimodal perception pipelines have also been explored in TinyML deployments. For example, lightweight sensor processing on multiple cores combined with specialised hardware accelerators such as TPUs can enable efficient real-time behaviour recognition directly on edge devices~\cite{Kanjo2026parallel} .

Multimodal fusion is therefore best viewed as a \emph{scheduling- and synchronization-dominated} workload. Individual models may satisfy latency targets in isolation yet fail once composed through shared middleware and memory. The resulting errors are not necessarily model errors; they are timing errors. This workload class directly motivates the synchronization, scheduling, and communication barriers that appear later in the Deployment Gauntlet.

\paragraph{Section summary.}
The workload classes in Table~\ref{tab:edge_workloads_a} motivate a systems taxonomy rather than a model-centric account of edge deployment. VLA policies are dominated by memory traffic and weight streaming, diffusion policies by sustained compute and thermal load, vision encoders by visual prefill and bursty ingestion, 3D perception by sparsity and irregular memory access, and multimodal fusion by scheduling and synchronization. Section~\ref{sec:gauntlet} builds on this distinction and organizes the resulting deployment failures through the Deployment Gauntlet.

\section{The Deployment Gauntlet}
\label{sec:gauntlet}

Section~\ref{sec:landscape} showed that embodied foundation model workloads differ sharply in their dominant execution bottlenecks. In deployment, however, these workloads rarely operate in isolation. Memory intensive VLA policies, compute-heavy diffusion controllers, burst-driven visual encoders, and synchronization sensitive fusion pipelines are typically co-located on the same embedded system-on-chip (SoC), where they compete for shared memory bandwidth, thermal headroom, power budget, and scheduling priority~\cite{bechtel2024analysis}. As a result, end-to-end failures often emerge not from any single model component, but from their interaction under shared physical constraints.

\begin{figure}[t!]  
   \centering
   \includegraphics[width=0.95\textwidth]{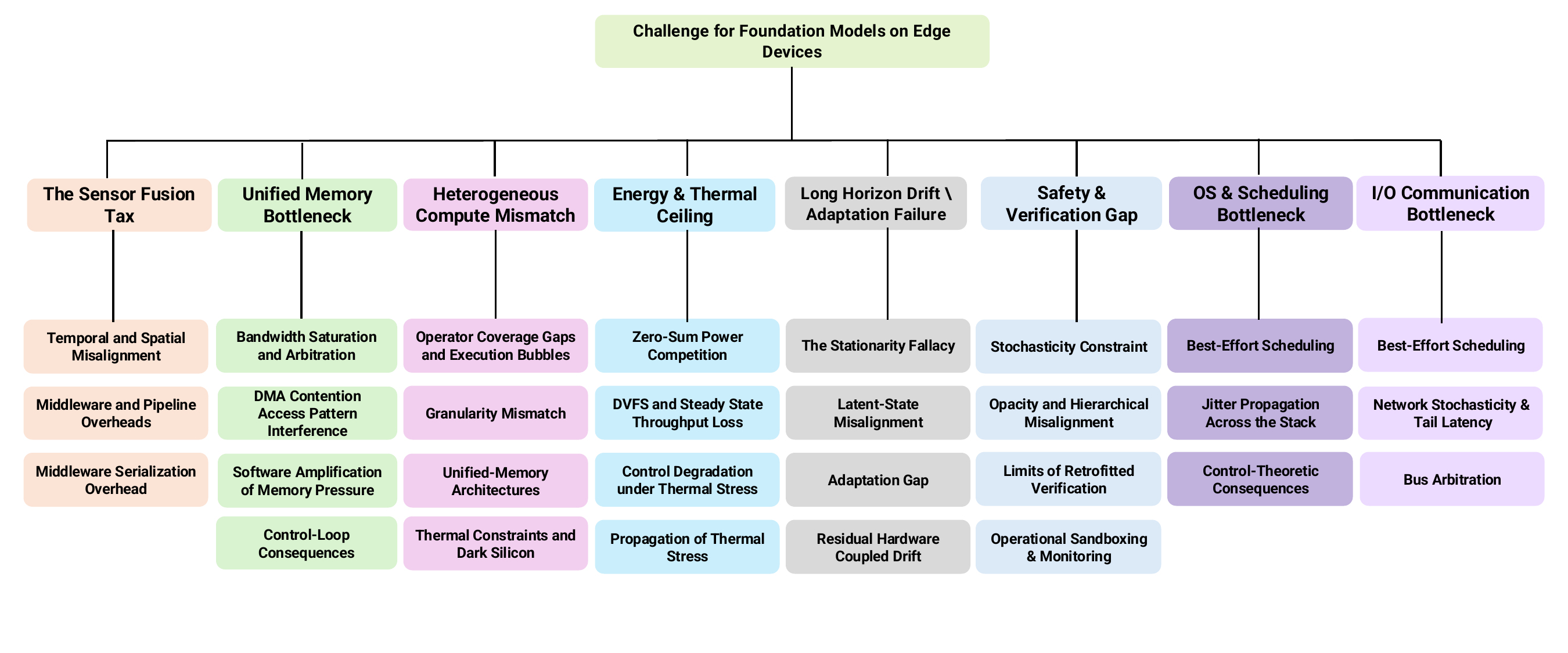} 
   \vspace{-6mm}
    \caption{The taxonomy for the gauntlets and solutions in Foundation Models under Embodied Constraints. }
    \vspace{-3mm}
    \label{fig:taxonomy-challenge-solution}
\end{figure}

This interaction is the basis of the \emph{Deployment Gauntlet}. We use the term to denote a systems taxonomy of the recurring barriers that embodied foundation models must satisfy to operate reliably on edge platforms. These barriers appear across platforms, modalities, and application domains~\cite{Firoozi2023FoundationMI} because they arise from a common deployment setting: closed-loop execution under tight constraints on memory, compute, I/O, timing, power, and safety. When these pressures compound, systems exhibit latency variance, deadline misses, synchronization failures, and thermal throttling, even when individual components remain tractable in isolation. 
Figure~\ref{fig:taxonomy-challenge-solution} summarizes the Deployment Gauntlet as a systems taxonomy of the eight coupled barriers that shape reliable embodied foundation model execution at the edge.

The Deployment Gauntlet is therefore not a list of isolated implementation defects, nor is it exhausted by local optimizations such as pruning, quantization, or accelerator specialization~\cite{hooker2020hardware}. Instead, it organizes the recurring systems constraints that emerge when embodied foundation models must sustain real-time perception, inference, and control on shared edge hardware. The remainder of this section develops the Gauntlet through eight coupled barriers, each corresponding to a distinct but strongly interacting source of deployment failure. Table~\ref{tab:gauntlet_challenges1_4} (Barriers I-IV) and Table~\ref{tab:gauntlet_challenges5_8} (Barriers V-VIII), that emerge repeatedly across platforms, modalities, and application domains \cite{Firoozi2023FoundationMI}

\subsection{The Sensor Fusion Tax}
\label{sub:sensor-fusion}

The \textbf{Sensor Fusion Tax} is the latency, jitter, and memory overhead incurred when heterogeneous, asynchronous sensor streams are transformed into time-aligned, control-ready representations. In embodied systems, this overhead can consume a substantial fraction of the control-cycle budget. It does not originate from a single bottleneck. Instead, it accumulates across temporal synchronization, spatial calibration, middleware orchestration, and representational translation. Foundation models intensify this burden because they add sustained memory traffic, increase sensitivity to timing error, and enlarge the execution graph that must remain stable under closed-loop operation.

\subsubsection{Temporal and Spatial Misalignment}
Robotic sensing is inherently asynchronous: cameras, LiDAR, and IMUs operate on different clock domains and exhibit variable transport delay. Enforcing temporal consistency therefore requires buffering and synchronization mechanisms such as ROS~2's \textit{ApproximateTime} filter. These improve alignment, but they also increase end-to-end latency and reduce control responsiveness. Li et al.~\cite{li2022worst} show that synchronization can idle downstream computation while waiting for delayed packets, reducing effective throughput. In high-speed manipulation, even millisecond-scale jitter can produce substantial divergence between perceived and true system state~\cite{jellum2022syncline}. Spatial fusion introduces a second source of overhead. Calibration degrades under vibration, thermal expansion, and mechanical drift, and corrective methods such as LCCNet~\cite{lv2021lccnet} and CalibRefine~\cite{cheng2025calibrefine} add nontrivial computation to the control loop. The resulting trade-off is straightforward: better alignment often consumes more of the available control-cycle budget.

\subsubsection{Middleware and Pipeline Overheads}
Middleware compounds these costs. ROS~2 enables modular message passing, but serialization and deserialization remain expensive even in intra-process settings, with reported latency penalties of up to 50\% relative to shared-memory designs~\cite{kronauer2021latency}. For high-bandwidth modalities such as LiDAR and 4K video, repeated copying consumes memory bandwidth that would otherwise be available to inference. Non-deterministic scheduling further creates queue buildup and jitter spikes on the order of 10--40\,ms~\cite{teper2024bridging}. Conventional fusion pipelines impose an additional constraint: they often block downstream inference until a complete sensor sweep has arrived, effectively tying control frequency to the slowest sensing modality. Pipelined systems such as MMEdge~\cite{huang2025mmedge} reduce average latency by overlapping sensing and inference, but they do not remove the underlying synchronization and scheduling burden.

\subsubsection{Representational Mismatch and Foundation Model Sensitivity}
Representational mismatch introduces a third source of overhead. Event cameras, for example, produce sparse, microsecond-resolution streams that do not map naturally to frame-based foundation models. Converting these streams into dense tensors through reconstruction or voxelization can offset part of the energy advantage of neuromorphic sensing and increase CPU-side preprocessing overhead when dedicated support is unavailable~\cite{gallego2020event,gehrig2020video}. Foundation models then amplify the fusion burden in three ways. First, sustained parameter movement can saturate LPDDR bandwidth on unified-memory systems and reduce headroom for sensor buffering and transport~\cite{bommasani2021opportunities}. Second, end-to-end VLA policies can be highly sensitive to temporal misalignment, so synchronization error propagates directly into action quality~\cite{tang2025e3ad}. Third, practical control rate is often bounded not by nominal inference throughput, but by the slowest stage in the sensing and synchronization path~\cite{ruan2025survey}.

\subsubsection{Implications}
The Sensor Fusion Tax is a recurring systems constraint in embodied deployment, not a narrow implementation artifact. Once perception is embedded in a real-time control stack, synchronization, calibration, and data movement become first-order determinants of latency stability, memory pressure, and closed-loop reliability.
\begin{table}[t]
\centering
\caption{Deployment Gauntlet barriers 1--4: recurring system-level challenges for embodied foundation models at the edge.}
\label{tab:gauntlet_challenges1_4}
\scriptsize
\setlength{\tabcolsep}{3.5pt}
\renewcommand{\arraystretch}{1.12}
\begin{tabularx}{\columnwidth}{|m{1.75cm}|m{2.55cm}|X|}
\hline
\rowcolor[HTML]{800000}
{\color{white}\textbf{Barrier}} &
{\color{white}\textbf{Representative  Subproblem}} &
{\color{white}\textbf{Systems Effect}} \\
\hline\hline

\multirow{3}{1.75cm}{Sensor Fusion Tax}
& Temporal and Spatial Misalignment
& Asynchronous sensor clocks, transport delay, and calibration drift force buffering, synchronization, and corrective alignment, trading control responsiveness for temporal and spatial consistency~\cite{li2022worst,jellum2022syncline,lv2021lccnet,cheng2025calibrefine}. \\
\cline{2-3}

& Middleware and Pipeline Overheads
& Serialization, copying, buffering, and sweep-level pipeline blocking consume bandwidth and introduce latency jitter that destabilizes high-frequency control execution even when average throughput is acceptable~\cite{kronauer2021latency,teper2024bridging,huang2025mmedge}. \\
\cline{2-3}

& Representational Mismatch and Foundation-Model Sensitivity
& Modalities such as event streams require costly reconstruction or voxelization before they can be processed by frame-based models, while foundation models further amplify fusion overhead through higher memory traffic and greater sensitivity to timing error~\cite{gallego2020event,gehrig2020video,bommasani2021opportunities,tang2025e3ad,ruan2025survey}. \\
\hline

\multirow{4}{1.75cm}{Heterogeneous Compute Mismatch}
& Operator Coverage Gaps and Execution Bubbles
& Limited operator support on edge accelerators forces graph partitioning, CPU fallback, and host--device synchronization, creating execution bubbles and large latency spikes that reduce end-to-end throughput~\cite{he2022empointmovseg,chen2025nanovla}. \\
\cline{2-3}

& Granularity Mismatch and Kernel Launch Overhead
& Fine-grained transformer kernels incur substantial launch overhead on embedded GPUs, accounting for 30--60\% of latency and driving utilization below 20\% during autoregressive decoding~\cite{zeng2024flightllm}. \\
\cline{2-3}

& The Memory Wall under Unified-Memory Architectures
& Shared LPDDR bandwidth contention among CPUs, GPUs, NPUs, and sensors makes execution memory-bound and can offset nominal hardware-acceleration gains through arbitration and tensor movement overhead~\cite{ivanov2021data,kim2024openvla}. \\
\cline{2-3}

& Thermal Constraints and Dark Silicon
& Sustained multimodal workloads trigger DVFS throttling, reduce steady-state throughput, and make nominal peak compute difficult to maintain during prolonged execution~\cite{karumbunathan2022nvidia,esmaeilzadehretrospective}. \\
\hline

\multirow{4}{1.75cm}{Unified Memory Bottleneck}
& Bandwidth Saturation and Arbitration
& Concurrent FM parameter streaming and high-rate sensor ingestion saturate shared LPDDR bandwidth, so accelerators stall because memory delivery, rather than arithmetic throughput, becomes the limiting factor~\cite{akkad2023embedded,shazeer2019fast,zeng2024flightllm}. \\
\cline{2-3}

& DMA Contention and Access Pattern Interference
& Sensor DMA traffic competes directly with accelerator memory access, while mismatched access patterns reduce cache efficiency and increase latency variance across perception and control loops~\cite{bechtel2024analysis,ivanov2021data}. \\
\cline{2-3}

& Software Amplification of Memory Pressure
& Middleware buffering, serialization, and large FM weight footprints amplify memory pressure, triggering reallocations, evictions, and other memory-management events that reduce throughput and destabilize execution~\cite{kim2024openvla}. \\
\cline{2-3}

& Control-Loop Consequences
& Memory stalls delay accelerators, block control threads, increase thermal load, and degrade real-time behavior before nominal compute limits are reached~\cite{teper2024bridging,karumbunathan2022nvidia,gholami2021survey}. \\
\hline

\multirow{5}{1.75cm}{Energy \& Thermal Ceiling}
& Zero-Sum Power Competition
& Foundation-model inference competes directly with propulsion, actuation, and sensing for a fixed onboard energy budget, reducing endurance under sustained execution~\cite{tan2024autonomous,zeng2024flightllm,horowitz20141}. \\
\cline{2-3}

& DVFS and Steady-State Throughput Loss
& High-duty-cycle workloads exceed passive cooling capacity and trigger DVFS transitions that reduce sustained throughput once the platform reaches thermal equilibrium~\cite{swaminathan2025benchmarking}. \\
\cline{2-3}

& Control Degradation under Thermal Stress
& Reduced steady-state inference rate increases perception staleness and lowers closed-loop responsiveness by slowing state updates and control decisions~\cite{swaminathan2025benchmarking}. \\
\cline{2-3}

& Propagation of Thermal Stress
& Thermal coupling across the SoC can throttle CPUs responsible for sensing, middleware, and control orchestration even when the model workload itself is unchanged~\cite{vanintroducing}. \\
\cline{2-3}

& Control-Loop Jitter and Modal Trade-offs
& Thermal limits can introduce actuation jitter and force the system to stagger, reduce, or selectively disable sensing modalities to remain within sustained power and thermal budgets~\cite{vanintroducing,chen2025nanovla,wen2025tinyvla,luo2023latent}. \\
\hline

\end{tabularx}
\end{table}
\FloatBarrier

\subsection{Heterogeneous Compute Mismatch}
\label{sec:compute_mismatch}

The \textbf{Heterogeneous Compute Mismatch} arises when embodied foundation model workloads are mapped onto edge platforms whose CPUs, GPUs, and NPUs are optimized for different classes of operators and execution regimes. Modern systems such as Jetson AGX Orin and Apple M-series devices rely on this heterogeneity to balance performance and energy efficiency under tight thermal limits~\cite{gill2025edge}. Embodied FM workloads, however, combine dense tensor kernels, irregular preprocessing, host-side coordination, and latency-sensitive control in the same execution graph. The result is a recurring systems barrier: hardware heterogeneity improves nominal efficiency, but it undermines predictable low-latency execution. In practice, latency becomes sensitive to operator placement, runtime fragmentation, and scheduling effects rather than to peak compute alone~\cite{hooker2020hardware}.

\subsubsection{Operator Coverage Gaps and Execution Bubbles}
Foundation-model inference depends on a broad operator set, including attention, normalization, tensor reshaping, and irregular preprocessing, whereas edge NPUs accelerate a much narrower subset of dense kernels such as INT8 matrix multiplication. Unsupported operators force graph partitioning and CPU fallback, introducing host--device synchronization and interrupting otherwise continuous execution. These transitions create \emph{execution bubbles} in which accelerators idle while control returns to the host. Sparse perception stages such as LiDAR voxelization are especially prone to these fallbacks and can introduce large latency spikes~\cite{he2022empointmovseg}. NanoVLA shows that removing even small CPU-resident operators can improve throughput by up to $1.7\times$, highlighting that fragmented execution, not peak compute alone, often determines end-to-end performance~\cite{chen2025nanovla}.

\subsubsection{Granularity Mismatch and Kernel Launch Overhead}
Foundation-model inference also mismatches the execution granularity of mobile GPUs. Transformer decoding decomposes into many fine-grained kernels, but embedded GPUs cannot amortize launch overhead as effectively as larger server-class devices. On Jetson Orin, CPU-side launch overhead can exceed kernel execution time and account for 30--60\% of end-to-end latency during autoregressive decoding~\cite{zeng2024flightllm}. Under these conditions, runtime overhead becomes a major component of inference cost. For VLA policies, sequential token generation can drive utilization below 20\%, making multi-billion-parameter models difficult to reconcile with high-frequency control.

\subsubsection{The Memory Wall under Unified-Memory Architectures}
Unified-memory architectures amplify the mismatch further. On embedded SoCs, CPUs, GPUs, NPUs, and sensor DMA engines share the same LPDDR channels, so FM inference competes directly with high-rate perception traffic for bandwidth~\cite{ivanov2021data}. Diffusion-based policies intensify this pressure by revisiting large tensors across repeated denoising steps and sustaining DRAM traffic over long intervals. Cross-device tensor migration introduces an additional penalty. Moving high-resolution features between heterogeneous units, such as the CPU and NPU, serializes execution and can add 4--15\,ms of latency, often offsetting the nominal benefit of hardware acceleration~\cite{kim2024openvla}. In this regime, end-to-end performance is constrained less by arithmetic throughput than by memory movement and bandwidth arbitration.

\subsubsection{Thermal Constraints and Dark Silicon}
Thermal limits turn these inefficiencies into sustained deployment failures. Multimodal embodied pipelines often activate CPUs, GPUs, and NPUs concurrently, pushing passive or compact cooling systems beyond their practical envelope. Dynamic voltage and frequency scaling (DVFS) responds by reducing clock rates, and mixed-workload saturation has been reported to reduce steady-state inference throughput by as much as 60\% relative to cold-start performance~\cite{karumbunathan2022nvidia}. Foundation-model workloads are especially exposed because they sustain high utilization over time rather than in short bursts. As a result, nominal peak performance becomes difficult to maintain during prolonged operation, and systems that initially satisfy real-time targets may later fall below them as thermal throttling accumulates~\cite{esmaeilzadehretrospective}.

\subsubsection{Implications}
Heterogeneous Compute Mismatch is therefore not just an efficiency loss; it is a stability problem for closed-loop embodied execution. Once inference spans heterogeneous units under shared thermal and memory limits, operator placement, launch overhead, and runtime fragmentation become first-order determinants of whether the system can sustain safe real-time behavior.
\subsection{The Unified Memory Bottleneck}
\label{sec:memory_bottleneck}

The \textbf{Unified Memory Bottleneck} arises when embodied workloads force inference, sensing, and runtime services to contend for the same LPDDR bandwidth and memory service time. Unlike server platforms with dedicated GPU VRAM, embedded SoCs place CPUs, GPUs, NPUs, and sensor peripherals on a shared memory fabric. Under these conditions, performance is constrained less by nominal capacity than by arbitration and data movement under concurrent load~\cite{akkad2023embedded}. In embodied systems, that contention directly degrades latency stability and closed-loop responsiveness.

\subsubsection{Bandwidth Saturation and Arbitration}
Foundation-model inference places sustained pressure on memory bandwidth. Autoregressive decoding and diffusion-based policies repeatedly stream large parameter tensors from DRAM~\cite{shazeer2019fast}. Data-center accelerators can mask much of this cost with bandwidth on the order of terabytes per second, whereas edge SoCs typically operate in the 25--204\,GB/s range. In embodied deployment, that same bandwidth must also serve RGB cameras, LiDAR, event streams, and other high-rate inputs. As a result, accelerators often fail to reach nominal arithmetic throughput because execution is limited by memory supply rather than compute availability~\cite{zeng2024flightllm}. As arbitration pressure increases, latency variance rises and real-time margins shrink.

\subsubsection{DMA Contention and Access Pattern Interference}
DMA traffic adds a second source of interference. On platforms such as Jetson AGX Orin, image signal processors, USB controllers, and other peripheral interfaces compete with GPU and NPU kernels for the same memory cycles~\cite{bechtel2024analysis}. Because DMA traffic is bursty and often hardware-prioritized, inference can stall even when average bandwidth appears sufficient. Access-pattern mismatch further reduces effective throughput. Linear sensor streams interfere with the poor-locality access patterns of attention layers and sparse LiDAR pipelines, reducing cache efficiency and degrading bandwidth utilization~\cite{ivanov2021data}. Under unified memory, this interference disrupts both sensor buffering and model weight reuse.

\subsubsection{Software Amplification of Memory Pressure}
Software abstractions magnify the hardware bottleneck. Middleware frameworks such as ROS~2 often serialize, duplicate, or buffer high-bandwidth sensor messages across node boundaries. When large models already consume 6--8\,GB for weights alone, these additional image and LiDAR buffers leave little headroom for stable execution. Small shifts in allocation demand can then trigger buffer reallocation, tensor rematerialization, page eviction, and other costly memory-management events. The result is latency jitter, throughput loss, and, in stressed regimes, node instability or watchdog resets~\cite{kim2024openvla}. The bottleneck therefore scales with model size not only because of hardware limits, but also because software layers amplify memory pressure.

\subsubsection{Control-Loop Consequences}
The effects of memory contention extend well beyond inference throughput. When DRAM service becomes unreliable, NPUs stall waiting for on-chip buffers to refill, and CPU threads responsible for synchronization and control queue behind memory-bound tasks. In robotic systems operating at 50--100\,Hz, these delays appear as actuator jitter, delayed servo response, and degraded sensor fusion~\cite{teper2024bridging}. Sustained DRAM traffic also raises power draw and accelerates thermal saturation, triggering DVFS before compute units are fully utilized~\cite{karumbunathan2022nvidia}. Empirical studies show that workloads such as diffusion policies can drop from roughly 30\,FPS in isolation to 6--12\,FPS under realistic sensor load because of memory contention alone. In embodied settings, this is not just a throughput penalty; it directly affects the freshness of observations presented to the control loop~\cite{gholami2021survey}.

\subsubsection{Implications}
The Unified Memory Bottleneck is therefore a first-order systems barrier in embodied edge deployment. Once sensing, inference, and control share the same memory fabric, bandwidth arbitration, DMA behavior, and software buffering become direct determinants of whether the platform can sustain safe real-time operation.
\begin{table}[t]
\centering
\caption{Deployment Gauntlet barriers 5--8: recurring system-level challenges for embodied foundation models at the edge.}
\label{tab:gauntlet_challenges5_8}
\scriptsize
\setlength{\tabcolsep}{3.5pt}
\renewcommand{\arraystretch}{1.12}
\begin{tabularx}{\columnwidth}{|m{1.72cm}|m{2.45cm}|X|}
\hline
\rowcolor[HTML]{800000}
{\color{white}\textbf{Barrier}} &
{\color{white}\textbf{Representative Subproblem}} &
{\color{white}\textbf{Systems Effect}} \\
\hline\hline

\multirow{4}{1.72cm}{Long-Horizon Execution Drift and Adaptation Failure}
& The Stationarity Fallacy
& Embodied deployment violates the stationarity assumptions of offline training: sensor noise, actuator behavior, and calibration drift shift over time, causing prediction error to accumulate as observations move away from well-supported regions of the training distribution~\cite{kumar2021rma}. \\
\cline{2-3}

& Latent-State Misalignment
& Physical drift can disrupt the alignment between sensory embeddings and action-relevant latent structure, causing internally coherent predictions to diverge from the true physical state of the platform~\cite{ha2018world}. \\
\cline{2-3}

& The Adaptation Gap
& Edge platforms often cannot support timely online adaptation under SWaP constraints, while cloud-mediated adaptation introduces latency and connectivity dependence, leaving long-duration deployments governed by increasingly outdated model assumptions~\cite{ren2025surfer,gao2025test}. \\
\cline{2-3}

& Residual Hardware and Environment-Coupled Drift
& Sensor bias instability, thermal noise, mechanical wear, and terrain-dependent slip introduce residual time-varying error that classical compensation can attenuate but not fully remove~\cite{zhou2025learning,vargas2021overview,sigron2023compensation,sakayori2024modeling}. \\
\hline

\multirow{4}{1.72cm}{Safety \& Verification Gap}
& Stochastic Objectives and Constraint Violations
& Likelihood-based objectives such as token prediction and diffusion denoising do not inherently encode hard physical constraints, so actions may be statistically plausible while remaining dynamically unsafe under real-world perturbations~\cite{tolle2025towards}. \\
\cline{2-3}

& Opacity and Hierarchical Misalignment
& High-dimensional latent representations are difficult to interpret formally, and mismatched interfaces between learned planners and constrained controllers can yield semantically valid but dynamically infeasible actions~\cite{Firoozi2023FoundationMI,Yang2023FoundationMF,Ahn2022DoAI}. \\
\cline{2-3}

& Limits of Retrofitted Verification
& Barrier certificates, reachability analysis, and fallback logic provide only partial coverage when state estimates are shifted, uncertainty is poorly calibrated, or policy dimensionality exceeds the scope of practical formal analysis. \\
\cline{2-3}

& Operational Sandboxing and Structural Implications
& Reduced velocities, geofencing, teleoperation, and runtime safety monitors lower operational risk, but they remain reactive and do not remove the underlying mismatch between generative model behavior and physically constrained control~\cite{sharifi2025system,mirzaeian2021diverse}. \\
\hline

\multirow{3}{1.72cm}{OS \& Scheduling Bottleneck}
& Best-Effort Scheduling and Runtime Nondeterminism
& Throughput-oriented schedulers and runtime-level serialization introduce jitter, descheduling, and callback delay that are difficult to reconcile with bounded-latency control execution~\cite{shen2025sentryrt,sobhani2023timing}. \\
\cline{2-3}

& Jitter Propagation Across the Stack
& Descheduling, priority inversion, and shared-resource contention propagate across sensing, inference, and actuation, producing bursty sensing, irregular frame timing, and cross-layer jitter cascades~\cite{sobhani2023timing}. \\
\cline{2-3}

& Control-Theoretic Consequences
& Best-effort timing violates assumptions such as bounded execution and regular sampling, introducing heavy-tailed delay that can manifest as oscillation, missed grasps, or delayed corrective action~\cite{gomes2000heavy}. \\
\hline

\multirow{2}{1.72cm}{I/O \& Communication Bottleneck}
& Bus Arbitration, Network Stochasticity, and Middleware Overhead
& Shared DMA engines, LPDDR arbitration, network delay, and middleware serialization jointly disrupt bounded-delay multimodal transport, degrading throughput and timing regularity under load~\cite{park2025real,navardi2025genai,maruyama2016exploring}. \\
\cline{2-3}

& Cross-Layer Timing Cascades
& Delayed data movement can stall inference, misalign fusion, reorder messages, and propagate timing inconsistency across perception, planning, and control rather than remaining local to the transport layer. \\
\hline

\end{tabularx}
\end{table}
\FloatBarrier
\vspace*{-.2cm}
\subsection{The Energy and Thermal Ceiling}
\label{sec:energy_thermal}

The \textbf{Energy and Thermal Ceiling} arises because mobile platforms must execute foundation models within fixed power, cooling, and endurance budgets. Unlike cloud infrastructure, mobile and aerial robots operate under strict Size, Weight, Power, and Cost (SWaP-C) constraints and have little thermal headroom. Under these conditions, sustained inference competes directly with propulsion, sensing, and actuation for both energy and heat dissipation capacity~\cite{tan2024autonomous}. As a result, embodied performance is often bounded by steady-state power and thermal limits before nominal model capacity is reached.

\subsubsection{Zero-Sum Power Competition}
On embedded platforms, computation and physical actuation draw from the same onboard energy reservoir. In aerial systems, allocating an additional 10--15\,W to accelerator-driven transformer inference can reduce flight endurance by several minutes~\cite{zeng2024flightllm}. The trade-off is particularly severe for foundation models because their cost is driven not only by arithmetic but also by DRAM access, which remains among the most energy-intensive operations in modern hardware~\cite{horowitz20141}. Sustained memory traffic therefore raises both power draw and heat generation, forcing a direct trade-off between model throughput and operational range.

\subsubsection{DVFS and Steady-State Throughput Loss}
As the platform approaches its thermal envelope, Dynamic Voltage and Frequency Scaling (DVFS) lowers clock rates to maintain safe operation. The result is a drop in sustained throughput that can differ markedly from cold-start performance. A vision encoder that begins near 30\,FPS, for example, may settle at a much lower rate after thermal throttling. Benchmark studies indicate that high-duty-cycle foundation-model workloads often enter this regime within minutes of continuous execution~\cite{swaminathan2025benchmarking}. For embodied deployment, the relevant metric is therefore not startup throughput, but the control rate the platform can maintain after reaching thermal equilibrium.

\subsubsection{Control Degradation under Thermal Stress}
This steady-state slowdown propagates directly into the control stack. A policy that initially runs near 20\,Hz may stabilize at a much lower frequency after throttling, increasing perception staleness and reducing closed-loop responsiveness. In embodied systems, timing margins are often narrow, so reduced inference rate leads directly to slower state updates and delayed control decisions. Thermal stress therefore alters not only throughput, but the temporal behavior of the system as a whole.

\subsubsection{Propagation of Thermal Stress}
Thermal effects are also coupled across the SoC. CPUs, GPUs, and NPUs typically share a die-level or package-level thermal envelope, so sustained GPU or NPU utilization can trigger CPU throttling even when the model workload itself is unchanged. This is consequential because the CPU continues to handle sensor drivers, middleware callbacks, and control orchestration. When CPU frequency drops, those services slow as well, increasing end-to-end latency outside the model proper~\cite{vanintroducing}.

\subsubsection{Control-Loop Jitter and Modal Trade-offs}
Once throttling spreads through the system, control-loop timing degrades. A nominal 100\,Hz loop can develop substantial actuation jitter when CPU-bound planning and coordination stages stretch from single-digit milliseconds to tens of milliseconds~\cite{vanintroducing}. Thermal limits also constrain multimodal execution. Running high-resolution vision, LiDAR fusion, and continuous audio concurrently can exceed the thermal design power of compact edge SoCs, forcing the system to reduce rates, stagger execution, or disable modalities selectively. In this setting, compact architectures such as NanoVLA~\cite{chen2025nanovla}, TinyVLA~\cite{wen2025tinyvla}, and OneDP~\cite{luo2023latent} are not just efficiency improvements; they are practical mechanisms for keeping multimodal inference within sustained thermal and power budgets.

\subsubsection{Implications}
The Energy and Thermal Ceiling is a sustained-execution barrier, not a short-burst benchmarking artifact. For embodied deployment, the relevant question is not whether a model can run briefly at peak speed, but whether the platform can maintain the required sensing, inference, and control rate over time without exhausting energy reserves or destabilizing timing.
\subsection{Long-Horizon Execution Drift and Adaptation Failure}
\label{sec:drift_adaptation}

The \textbf{Long-Horizon Execution Drift and Adaptation Failure} barrier arises because embodied foundation models are deployed with mostly fixed representations in physical systems whose sensing, actuation, and environment change over time. Unlike cloud inference, embodied execution is inherently non-stationary: sensor characteristics shift, actuator behavior changes, and contact dynamics evolve across deployment horizons. When frozen models continue to assume a stable observation--action mapping under these conditions, prediction error accumulates and control quality degrades~\cite{sunderhauf2018limits}. The core difficulty is not drift alone, but drift combined with limited capacity for online correction on edge platforms.

\subsubsection{The Stationarity Fallacy}
Embodied deployment violates the stationarity assumptions embedded in offline training. Sensor noise changes with temperature, actuator output degrades with battery discharge, and calibration shifts under vibration and wear. These effects induce persistent covariate shift in $P(O_t)$. Classical robotics pipelines address such variation through online state estimation and model-based correction, but frozen foundation models typically lack an equally direct mechanism for updating their internal representation of routine physical change. As drift accumulates, observations move away from well-supported regions of the training distribution, and prediction error compounds over time~\cite{kumar2021rma}. This yields a gradual degradation mode that is often underrepresented in short-horizon evaluation.

\subsubsection{Latent-State Misalignment}
Long-horizon drift is especially problematic for policies that rely on latent state integration, including Vision-Language-Action (VLA) and diffusion-based controllers. These systems assume that sensory embeddings remain aligned with the action-relevant structure learned during training. Physical perturbations can disrupt that alignment, mapping observations into latent regions that are less informative for control. The model may then continue to produce internally coherent predictions while diverging from the true physical state of the platform~\cite{ha2018world}. Recovery becomes difficult because the same drift that corrupted the latent state also weakens the corrective value of subsequent observations.

\subsubsection{The Adaptation Gap}
The barrier is compounded by an \textbf{adaptation gap} at the edge. Online gradient-based adaptation is often incompatible with the compute, timing, and energy budgets of SWaP-constrained platforms, while cloud-mediated adaptation introduces latency and connectivity dependence. As a result, long-duration deployments may be governed by increasingly outdated model assumptions. Test-time adaptation methods are promising~\cite{ren2025surfer,gao2025test}, but current approaches often remain too costly or too slow for real-time edge execution. In practice, model error and physical drift can therefore reinforce one another over long horizons.

\subsubsection{Residual Hardware and Environment-Coupled Drift}
Not all long-horizon drift can be removed through classical compensation. Sensor and actuator behavior contain residual, time-varying effects that violate idealized estimator assumptions and remain only partially observable during deployment. At the sensor level, IMUs exhibit bias instability under temperature and vibration, establishing a persistent noise floor~\cite{zhou2025learning}. Vision and LiDAR sensors show related non-stationarity through thermally induced noise, gain variation, and dark-current shifts~\cite{vargas2021overview}. At the actuation level, mechanical wear changes friction and compliance, while terrain-dependent slip breaks rigid-body assumptions and accumulates odometric error~\cite{sigron2023compensation,sakayori2024modeling}. Classical filters can attenuate part of this drift, but residual error remains and continues to perturb the observation action mapping presented to the model.

\subsubsection{Implications}
Long-horizon execution drift is therefore not only a perception or calibration issue. It is a systems-level barrier that emerges when fixed model representations are coupled to evolving physical platforms without a practical mechanism for continual correction. Over long missions, robust deployment depends not just on initial model accuracy, but on whether the system can preserve alignment among sensing, latent state, and control as conditions change.
\subsection{The Safety and Verification Gap}
\label{sec:safety_verification}

The \textbf{Safety and Verification Gap} arises because foundation-model objectives, representations, and inference behavior do not provide the same safety assurances as explicitly constrained control pipelines. Classical control stacks are often built around known dynamics, feasibility constraints, and analyzable stability conditions; embodied foundation models instead inherit the stochasticity and opacity of large neural architectures. In these systems, strong predictive or generative performance does not imply constraint satisfaction or safe closed-loop behavior. The gap therefore reflects a structural mismatch between how these models are trained and what physical autonomy requires.

\subsubsection{Stochastic Objectives and Constraint Violations}
Foundation models are typically optimized for predictive likelihood, denoising quality, or sequence modeling objectives rather than explicit physical risk. Vision-Language-Action token prediction, diffusion denoising, and latent interpolation do not inherently encode hard constraints such as joint limits, contact stability, or collision margins. As a result, generated actions may be statistically plausible while still being dynamically unsafe. Modest perturbations, including sensor noise, lighting variation, or adversarial artifacts, can induce action errors and constraint violations in embodied settings~\cite{tolle2025towards}. The core limitation is straightforward: likelihood-based objectives do not, by themselves, enforce physical invariants at inference time.

\subsubsection{Opacity and Hierarchical Misalignment}
The verification problem is compounded by model opacity. Classical controllers expose interpretable state variables, local sensitivities, and feasibility structure; foundation models instead operate through high-dimensional latent representations whose relation to physical state is often indirect~\cite{Firoozi2023FoundationMI}. As a result, failure analysis depends heavily on empirical testing rather than formal inspection~\cite{Yang2023FoundationMF}. Opacity also complicates hierarchical integration. High-level semantic planners may lack explicit contracts with lower-level kinematic or feedback controllers, so a planner can produce linguistically valid but dynamically infeasible actions that downstream systems must reject or repair~\cite{Ahn2022DoAI}. The barrier is therefore not only model uncertainty, but also interface mismatch between learned planning and constrained control.

\subsubsection{Limits of Retrofitted Verification}
Retrofitting formal guarantees onto foundation-model policies remains difficult. Control Barrier Functions can enforce local safety conditions, but those guarantees are only as reliable as the state estimate and model assumptions on which they depend. Under hallucinated or shifted observations, those assumptions may no longer hold. Formal reachability analysis faces a different limitation: it does not scale easily to high-dimensional, stochastic, multimodal transformer-based policies. Weak uncertainty calibration further narrows the available safety margin. When a model cannot reliably express epistemic uncertainty, downstream systems have limited ability to reject low-confidence actions or trigger fallback behavior before error accumulates. In practice, retrofitted verification remains useful, but it covers only part of the safety problem and degrades under the same distribution shifts that challenge the policy itself.

\subsubsection{Operational Sandboxing and Structural Implications}
Current embodied deployments often manage this gap through operational containment: reduced velocities, geofencing, supervised execution, and human teleoperation oversight. Runtime safety monitors that predict and interrupt imminent violations provide an important additional layer of protection~\cite{sharifi2025system}. However, these mechanisms remain largely reactive and depend on the same sensing and representation stack as the policy they supervise. They reduce operational risk, but they do not remove the underlying mismatch between generative model behavior and physically constrained control. Progress will require tighter integration of physical invariants, constraint satisfaction, and uncertainty estimation into the learning and control architecture itself~\cite{mirzaeian2021diverse}.

\subsubsection{Implications}
The Safety and Verification Gap is therefore a systems-level barrier, not merely a control-theoretic one. In embodied deployment, the central question is not whether a model can generate a plausible action, but whether the full stack can establish that the action is feasible, safe, and reliable under uncertainty before it reaches the plant.
\subsection{The OS and Scheduling Bottleneck}
\label{sec:os_scheduling}

The \textbf{OS and Scheduling Bottleneck} arises because embodied foundation-model pipelines often execute on software stacks optimized for flexibility and throughput rather than bounded latency. Classical robotic control systems commonly rely on RTOS or bare-metal execution to enforce explicit timing guarantees, whereas FM-based embodied stacks are often built on Linux kernels, containerized environments, and Python-centered orchestration. These layers simplify integration and support heterogeneous workloads, but they also introduce timing variability across kernel scheduling, middleware execution, and user-space coordination. In closed-loop embodied systems, that variability weakens the timing guarantees on which safe control depends.

\subsubsection{Best-Effort Scheduling and Runtime Nondeterminism}

A primary source of instability is Linux's Completely Fair Scheduler (CFS), which is designed to maximize aggregate throughput rather than worst-case latency. Under this policy, safety-critical control loops compete with logging, networking, thermal management, and other auxiliary processes for CPU time. At control rates of roughly 100--500\,Hz, even modest descheduling becomes consequential. On Jetson-class platforms, empirical studies report jitter spikes of 5--20\,ms caused by kernel wake-ups, interrupt bursts, and I/O-driven preemption~\cite{shen2025sentryrt}. Once scheduling delay approaches or exceeds the control period, the problem is no longer throughput; it is a direct loss of timing integrity in the control loop.

Python-based orchestration compounds the same issue. The Global Interpreter Lock serializes concurrent execution, and garbage collection can introduce pauses that are both unbounded and opaque to the scheduler. As a result, common coordination mechanisms in perception and control stacks, including asynchronous coroutines and ROS~2 executors, are difficult to reconcile with predictable real-time execution. Callback delays of only 5--10\,ms can disrupt timestamp alignment, propagate stale observations into Vision-Language-Action policies, and increase execution drift and collision risk~\cite{sobhani2023timing}. These effects are not isolated anomalies; they follow from the interaction between best-effort scheduling and runtime-level serialization.

\subsubsection{Jitter Propagation Across the Stack}

Multimodal workloads amplify this problem through priority inversion and shared-resource contention. Background LiDAR decoding threads may preempt safety monitors, asynchronous data loaders may delay control execution, and ROS~2 executors may defer perception updates while servicing unrelated timers~\cite{sobhani2023timing}. The immediate effect is bursty sensing, irregular frame timing, and transient sensor dropouts.

The more serious issue is that latency does not remain local to the component in which it originates. CPU descheduling can delay GPU kernel dispatch, and DMA activity from one sensor can block service to another. A single LiDAR burst may therefore delay camera buffering and trigger a \emph{jitter cascade} across perception, planning, and actuation. Under these conditions, learned policies and predictive filters must operate on temporally inconsistent inputs rather than merely noisy ones.

\subsubsection{Control-Theoretic Consequences}

OS-level nondeterminism undermines several assumptions used in classical control analysis. Stability arguments typically depend on bounded execution times, Zero-Order Hold discretization, and noise models that do not exhibit heavy-tailed latency behavior~\cite{gomes2000heavy}. Best-effort scheduling violates those assumptions by introducing irregular sampling and actuation delay. In practice, the resulting failures may appear as oscillations, missed grasps, or delayed corrective action and are easily misattributed to perception or policy quality, even when the underlying cause is temporal interference in the execution stack.

\subsubsection{Implications}

The OS and Scheduling Bottleneck is therefore a systems barrier, not merely an implementation detail. Gains in model efficiency or accelerator utilization do not by themselves restore bounded-latency execution on best-effort software stacks. Embodied FM systems that require dependable physical interaction need real-time-aware runtimes, priority-sensitive scheduling, and explicit timing contracts across sensing, inference, and control.
\subsection{The I/O and Communication Bottleneck}
\label{sec:io_communication}

The \textbf{I/O and Communication Bottleneck} arises because embodied foundation-model pipelines depend on multimodal data arriving with bounded delay, consistent ordering, and usable cross-modal synchronization, yet the underlying I/O stack cannot reliably preserve all three under load. Closed-loop perception and control often operate on sub-$10$\,ms timing budgets, but those budgets are eroded by arbitration in the I/O hierarchy, network variability, and middleware overhead. As a result, failures often originate not in model inference alone, but in the data path that feeds and coordinates it.

\subsubsection{Bus Arbitration, Network Stochasticity, and Middleware Overhead}

The first source of instability appears in the on-chip interconnect and memory subsystem. High-resolution RGB, high-rate IMU, and dense LiDAR streams compete for shared DMA engines and LPDDR bandwidth, which must also support foundation-model inference. This creates direct coupling between sensing and inference: when I/O pressure rises, arbitration delays data movement for both. On Orin-class platforms, such interference has been reported to reduce Vision-Language-Action throughput by 15--30\% under background I/O load, primarily because of arbitration stalls rather than compute saturation~\cite{park2025real}.

In cloud-assisted and split-compute deployments, the same problem extends beyond the device. Packet loss, retransmissions, fading, and queueing introduce heavy-tailed delay, and round-trip latencies of $30$--$80$\,ms on Wi-Fi~6 can exceed the stability margins of high-frequency control. Related communication limits also affect federated and fleet learning, where aggressive compression and update throttling are often required to keep communication overhead manageable~\cite{navardi2025genai}. The main issue is therefore not bandwidth alone, but the inability to maintain regular timing across sensing, inference, and remote coordination.

Middleware adds a further source of timing distortion. ROS~2 and DDS introduce serialization, copying, marshalling, and discovery overheads that consume CPU time and memory bandwidth for large tensors~\cite{maruyama2016exploring}. More importantly, DDS QoS policies reshape delivery timing through buffering, queueing, and retransmission, which can disrupt cross-modal simultaneity. Without zero-copy transport and explicit end-to-end timing contracts, middleware can invalidate the temporal assumptions required for reliable fusion and state estimation.

\subsubsection{Cross-Layer Timing Cascades}
These timing disturbances do not remain local to the transport layer. Memory contention can delay DMA, delayed DMA can stall inference, stalled inference can misalign fusion, and middleware buffering can reorder or defer messages. Thermal throttling may further slow packet processing and callback execution. The result is degraded temporal consistency across perception, planning, and control.

In embodied systems, that inconsistency has direct operational consequences. Obstacle updates become stale, fused state estimates lose coherence, and control actions are issued against an environment that has already changed. These failures are not purely algorithmic; they emerge from the interaction between model execution and the I/O path that supplies its inputs.

\subsubsection{Implications}
The I/O and Communication Bottleneck is therefore a systems barrier rooted in data movement, ordering, and timing, not just in raw compute latency. Mitigating it requires data-centric redesign, including prioritization of safety-critical traffic, near-sensor preprocessing and compression, and zero-copy communication paths for large tensor flows~\cite{shi2020communication}. Without such changes, improvements in model efficiency alone will not eliminate timing failures in embodied foundation-model deployment.
\section{Mitigation Families for the Deployment Gauntlet}
\label{sec:mitigations}

\begin{figure}[t!]  
   \centering
   \includegraphics[width=0.95\textwidth]{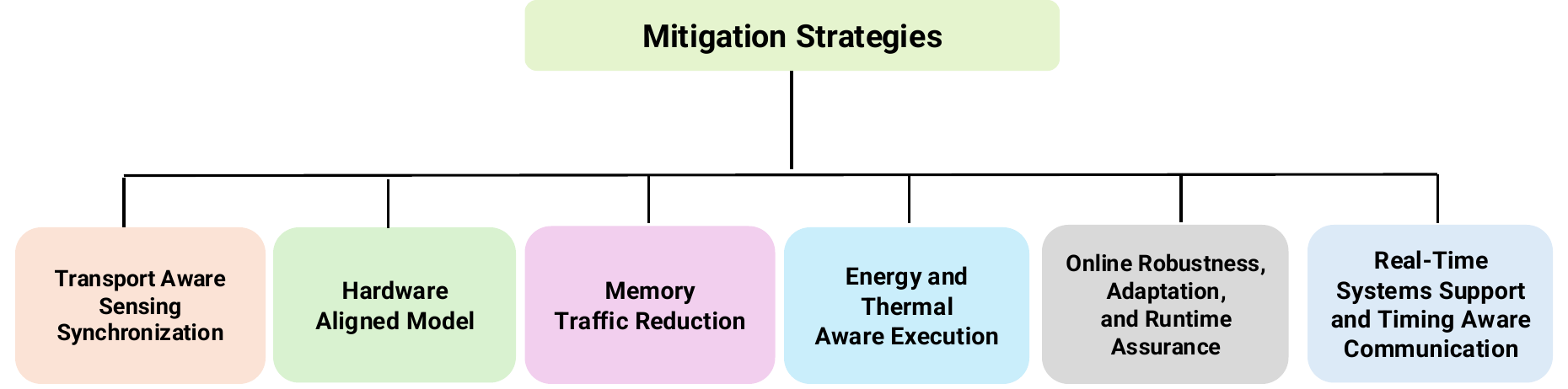} 
   \vspace{-3mm}
    \caption{The taxonomy for Mitigation Strategies for Deployment Gauntlet on edge }
    \vspace{-5mm}
    \label{fig:taxonomy-solution}
\end{figure}

The literature does not respond to the Deployment Gauntlet through eight isolated solution streams. The same mitigation often relieves multiple barriers at once by reducing synchronization cost, lowering memory traffic, or improving runtime predictability. For that reason, the solution space is better understood as a small set of recurring mitigation families rather than as a one-to-one mapping from barrier to fix. We group the literature into four such families: transport-aware sensing and synchronization relaxation, hardware-aligned model and compiler co-design, memory-traffic reduction and schedulable memory use, and energy- and thermal-aware execution. Figure~\ref{fig:taxonomy-solution} summarizes the Mitigation strategies for deployment gauntlet

\subsection{Transport-Aware Sensing and Synchronization Relaxation}

The first family acts on the data path before model inference begins. Its goal is to reduce transport overhead, relax unnecessary synchronization, and preserve temporal structure without forcing the full stack to wait on the slowest modality. Early work pursued this through zero-copy middleware transport. Iceoryx introduced shared-memory communication for fixed-size messages, while ROS~2 formalized loaned messages for middleware-managed buffer reuse~\cite{ishikawaaso2025agnocast,ros2zerocopy_design,ishikawaaso2025agnocast}. Profiling studies show that default ROS~2 configurations can incur substantial latency relative to optimized shared-memory pipelines, with behavior strongly shaped by DDS and executor design~\cite{kronauer2021latency}. Executor-level refinements were therefore introduced to reduce jitter and improve predictability under real-time load~\cite{abaza2024jitter,teper2024bridging,sobhani2023timing}.

Recent work extends this idea beyond fixed-size message transport. Agnocast generalizes zero-copy communication to variable-sized autonomy messages while preserving ROS~2 compatibility~\cite{ishikawaaso2025agnocast}. NITROS maintains accelerator-friendly formats through type adaptation and reduces unnecessary copies for high-resolution sensor streams~\cite{nitros2025}. At the fusion level, MMEdge reduces idle waiting through pipelined sensing and cross-modal speculative skipping~\cite{huang2025mmedge}, while AsyncVLA relaxes rigid action timing by generating control tokens asynchronously~\cite{jiang2025asyncvla}. Calibration maintenance belongs in the same family: CalibRefine lowers the downstream cost of spatial misalignment through targetless online refinement~\cite{cheng2025calibrefine}.

A related line of work reduces the need for strict synchronization altogether. Lightweight fusion modules, cooperative edge-oriented VLM pipelines, and hardware-aware multimodal co-design all aim to reduce how much data must be aligned exactly at runtime~\cite{yang2025edgebasedmultimodalsensordata,zhang2025miotsurfacevehicles}. Uncertainty-aware fusion methods dynamically gate modality contributions instead of assuming equal reliability and perfect simultaneity~\cite{lou2023uncertaintyencoded,cho2024cocoon,park2025resilientsensorfusionadverse}. Condition-aware pipelines similarly adapt to environmental regime rather than enforcing expensive redundancy~\cite{conditionaware2024multimodal}. Training can also absorb some of the burden. TimeAlign, UniBEV, and sensor-dropout strategies explicitly expose models to temporal misalignment or partial sensing, trading modest training complexity for substantially greater runtime slack~\cite{song2024timealignmultimodalobjectdetection,wang2024unibevmultimodal3dobject,liu2017learning,wang2025robustadversarialfusiontransformer}. Taken together, these methods treat sensing and synchronization as a transport-and-timing problem, not only as a perception problem.

\subsection{Hardware-Aligned Model and Compiler Co-Design}

The second family addresses the mismatch between transformer-style workloads and heterogeneous edge accelerators. The central idea is to expose hardware constraints at compile time or redesign operators and models so that execution remains on the accelerator as much as possible~\cite{wali2024hardware}. Compiler-driven approaches are the most direct expression of this idea. MATCH extends TVM-style compilation with explicit SoC hardware models and BYOC integration to optimize placement and scheduling across heterogeneous engines~\cite{hamdi2024matchmodelawaretvmbasedcompilation,gross2020hardware,lv2025duodecoding}. TinyIREE similarly unifies compilation and deployment for embedded targets and reduces execution fragmentation across devices~\cite{liu2022tinyireemlexecutionenvironment,rognlien2022hardware}. The shared lesson is that heterogeneity must be visible at compile time if host fallback and cross-engine synchronization are to be minimized.

When native hardware support is missing, operator reformulation becomes the next lever. T-MAN replaces unsupported low-bit primitives on NPUs with lookup-table abstractions that preserve end-to-end accelerator execution~\cite{wei2025tmanenablingendtoendlowbit,park2025mobilerag}. Quantization methods such as LLM.int8, GPTQ, and AWQ reduce tensor movement while preserving fidelity, improving accelerator residency and reducing host-device traffic~\cite{dettmers2022llmint88bitmatrixmultiplication,frantar2023gptqaccurateposttrainingquantization,lin2024awqactivationawareweightquantization}. Memory-aware attention redesigns such as FlashAttention reduce high-bandwidth memory access through IO-aware tiling~\cite{dao2022flashattentionfastmemoryefficientexact,dao2023flashattention,gupta2023flurka}, while FlightLLM targets host-device synchronization and kernel-launch overhead during autoregressive decoding~\cite{kwon2023efficientmemorymanagementlarge,zeng2024flightllm}.

The same design logic now appears at the architectural level. NanoVLA and TinyVLA redesign VLA pipelines for edge hardware by reducing redundant computation, sustaining accelerator residency, and separating heavyweight semantic reasoning from lightweight control~\cite{chen2025nanovla,smith2025flexigpt,wen2025tinyvla}. SARA-RT replaces quadratic attention with linear-complexity alternatives to recover real-time feasibility in transformer-based robotic control~\cite{leal2023sarartscalingroboticstransformers}. Across these approaches, robust edge deployment emerges less from post-training cleanup than from joint redesign of compiler, operator, and model.

\subsection{Memory-Traffic Reduction and Schedulable Memory Use}

The third family targets shared memory directly. On edge SoCs, the main problem is often not arithmetic throughput but whether LPDDR service remains predictable when foundation models, sensing, and control all compete for the same memory fabric. One response is isolation and regulation. Early work showed that shared-cache and DRAM interference can inflate tail latency even when average throughput remains acceptable~\cite{valsan2017addressing,bechtel2024analysis}. MemGuard-style reservation and policing mechanisms allocate DRAM budgets and throttle interfering workloads~\cite{yun2013memguard,7784697}. Later work extends these ideas to heterogeneous CPU--GPU systems through reserved bandwidth for real-time GPU tasks, accelerator-aware isolation, per-bank regulation, LLC control, and microsecond-scale policing~\cite{9211564,fang2024rt,puente2025rosguard,sullivan2024per,zuepke2024mempol}. In this line of work, memory is treated as a schedulable shared resource rather than a passive substrate.

A complementary strategy reduces traffic volume. Low-bit quantization lowers bandwidth demand while preserving accuracy; LifeQuant and ProxQ focus specifically on maintaining quantized performance in streaming or non-stationary settings~\cite{chen2023lifequant,cai2020zeroq,lin2020hrank}. Structured pruning reduces both parameter count and activation traffic through hardware-aware channel and block removal~\cite{li2023filter,liu2019metapruning,lin2019towards}. Rotation-based methods suppress activation outliers that would otherwise require precision escalation, and QuaRot extends low-bit execution across weights, activations, and KV caches~\cite{ashkboos2024slicegpt,chen2025efficientqat,ashkboos2024quarot,liu2024spinquant}.

The KV cache has become a particularly important target because it dominates long-context memory footprint. KIVI and related methods compress the cache through ultra-low-bit or coupled quantization~\cite{liu2024kivi,zhang2024kvcache1bit}, while adaptive-precision schemes vary cache precision across layers or tokens to allocate bits where they are most useful~\cite{hooper2025kvquant10millioncontext,shutova2025cachemustadaptivekeyvalue}. XQuant instead stores quantized layer inputs and rematerializes keys and values on demand, trading some on-chip computation for lower DRAM pressure~\cite{tomar2025xquant}. Across these approaches, the picture is consistent: once sensing and inference share unified memory, real-time viability depends on both lowering traffic volume and making the remaining traffic schedulable.

\subsection{Energy and Thermal-Aware Execution}

The fourth family addresses the fact that embodied deployment is governed by steady-state energy and thermal limits rather than by short-burst benchmark throughput. MAVBench makes this explicit by showing that autonomy workloads directly reduce vehicle endurance in micro aerial systems~\cite{boroujerdian2018mavbench}. At a lower level, Horowitz and later efficient-accelerator studies show that memory access often dominates arithmetic energy, which helps explain why memory-heavy foundation models generate disproportionate thermal load even when compute is optimized~\cite{horowitz20141,chen2016eyeriss,koppula2019eden}. The relevant design target is therefore not peak throughput, but \emph{joules per decision} and sustainable duty cycle under continuous execution~\cite{lee2021thermal,lin2023workload}.

One response lowers the cost of each inference step. Distillation and compact architectures such as MobileBERT reduce deployment cost while preserving useful capacity~\cite{sun2020mobilebert}. Post-training quantization methods including ZeroQuant and SmoothQuant further reduce compute and memory-transfer energy by lowering precision without large accuracy loss~\cite{yao2022zeroquant,xiao2023smoothquant}. These methods do not remove the thermal ceiling, but they shift the steady-state operating point.

For generative control, runtime duration matters as much as instantaneous cost. Diffusion-based policies are particularly demanding because repeated denoising accumulates heat over many steps~\cite{chi2025diffusion}. OneDP reduces that burden by collapsing multi-step diffusion into a single-step generator~\cite{wang2024one}, while LightDP prunes and restructures diffusion modules for sustained on-device execution~\cite{wu2025device}. In this regime, fewer generation steps directly improve thermal sustainability.

Thermal-aware runtime management provides the final lever. POD TAS uses predictive thermal-aware scheduling based on reduced-order thermal models to reduce peak temperature and temperature variance~\cite{dowling2023podtas,jacob2025thermml}. The broader lesson across this family is that thermal feasibility is a runtime property: compression lowers cost per operation, generative redesign reduces sustained duty cycle, and predictive scheduling shapes temperature growth over time. Under continuous edge deployment, sustainable embodied performance is determined by thermal equilibrium rather than by peak capability.

\subsection{Online Robustness, Adaptation, and Runtime Assurance}

Mitigations for long-horizon drift and safety failures share a common constraint: they must improve robustness without exceeding edge budgets for compute, timing, and energy. As a result, the literature does not pursue unrestricted online relearning of the full model. It instead relies on bounded corrective mechanisms that contain model mismatch before it propagates into unsafe closed-loop behavior. This family includes estimator-level drift compensation, parameter-efficient or gradient-free adaptation, control-theoretic safety filters, runtime assurance architectures, and reactive monitoring.

One strategy localizes correction outside the full perception or control backbone. In visual-inertial navigation, external neural modules estimate bias dynamics and inject corrections into factor-graph or invariant filtering pipelines~\cite{buchanan2022deep,altawaitan2025learned}. In soft robotics, where sensing drift is intrinsic to the hardware, continual learning with rehearsal~\cite{kushawaha2025adaptive} and RNN-based observers~\cite{george2022closing} track nonlinear drift while keeping compensation decoupled from task-level policy representations. These methods are effective precisely because they restrict adaptation to a narrow interface rather than perturbing the full model.

A second strategy constrains adaptation itself. Test-time adaptation methods suppress low-information gradients and stabilize streaming updates~\cite{Niu2022EfficientTM,Niu2023TowardsST,Gong2022NOTERC}, but on edge hardware such updates remain subordinate to latency-critical inference. Parameter-efficient methods narrow the update space further through sparse backpropagation~\cite{Zhu2023PockEngineSA}, lightweight bias modules~\cite{Cai2020TinyTLRM}, and bias-only transformer adaptation with well under $0.1\%$ of parameters~\cite{BenZaken2021BitFitSP}. When backpropagation is too costly or poorly supported by the hardware stack, gradient-free methods estimate updates from parameter perturbations~\cite{Niu2024TestTimeMA}, while meta-learning shifts most of the adaptation burden offline so that deployment requires only low-dimensional coefficient updates, as in Neural Fly~\cite{OConnell2022NeuralFlyER}. These approaches make adaptation feasible, but only within a deliberately limited correction range.

The same logic appears on the safety side. Control Barrier Function wrappers and related quadratic-program filters impose forward-invariant constraints around learned actions~\cite{ames2019control}, while Predictive Safety Filters and MPC-based supervisory schemes accept or replace actions according to an explicit feasibility model~\cite{wabersich2021predictive,tearle2021predictive}. Reachability-based intervention and RL-CBF variants similarly restrict unsafe exploration through safe sets or uncertainty-aware safety envelopes~\cite{fisac2018general,cheng2019end}, and SafeDiffuser extends this principle to diffusion-based planners by enforcing constraint-aware denoising~\cite{xiao2023safediffuser}. These methods improve safety by constraining or correcting learned behavior externally rather than by making the generative model intrinsically safe.

Runtime assurance architectures generalize that externalization. Simplex-style designs, SOTER, and related supervisory frameworks treat the learned policy as an advanced but uncertified controller that may be overridden by a verified baseline when safety is threatened~\cite{desai2019soter,chen2022runtime}. Formal verification methods such as Reluplex, Marabou, and neural reachability analysis provide stronger guarantees for small or structured networks~\cite{katz2017reluplex,katz2019marabou,ruan2018reachability}, but they do not scale naturally to high-dimensional multimodal foundation models under deployment shift. Probabilistic safety certificates, conformal wrappers, and runtime shielding partially address that limitation by combining risk-bounded guarantees, runtime enforcement, and black-box monitoring~\cite{tayal2025cp,zhang2025conformal,jansen2020safe,corsi2024verification,zolfagharian2024smarla}. In practice, these methods reduce operational risk, but they still depend on bounded correction, fallback, and supervision.

The central lesson is consistent across this family. Long-horizon drift and safety failures are mitigated through structured correction, constrained adaptation, and runtime oversight rather than through open-ended online relearning. These mechanisms improve robustness under realistic edge constraints, but they also make clear that dependable embodied deployment still depends on keeping correction bounded, explicit, and computationally tractable.

\subsection{Real-Time Systems Support and Timing-Aware Communication}

Mitigations for the OS, scheduling, I/O, and communication barriers address a different systems problem: preserving timing integrity across the execution stack. Their shared objective is to bound interference, reduce unnecessary data movement, and prevent timing disturbances from cascading across perception, planning, and control. Unlike model-centric optimizations, these methods operate on middleware, runtime, kernel, and transport layers.

At the middleware layer, chain-aware execution and callback isolation are central. Casini et al.~\cite{casini2019response} show that default ROS~2 executors produce pessimistic response-time bounds because of priority inversion and callback interference. Later work mitigates this by separating safety-critical callbacks from background activity~\cite{Yang2023FoundationMF} and by scheduling end-to-end execution chains rather than isolated callbacks. PiCAS is a representative example: it uses priority-driven, chain-aware scheduling to reduce perception-to-control latency substantially on embedded platforms~\cite{choi2021picas}. These methods improve timing predictability, but they remain dependent on lower-level scheduling and resource isolation.

Kernel-level real-time support addresses a second part of the same problem. PREEMPT\_RT substantially reduces CPU wake-up latency and improves interrupt handling under load~\cite{gutierrez2018real}, which is valuable for classical control and for the CPU-resident orchestration surrounding learned policies. Heterogeneous embodied systems, however, also depend on GPU kernels, DMA engines, and shared memory fabrics that the CPU scheduler governs only indirectly. OS-level real-time support therefore improves timing behavior but does not by itself guarantee end-to-end stability.

Runtime restructuring provides a more aggressive response. Systems such as Dora remove critical execution from Python and place it in compiled dataflow runtimes with zero-copy shared memory and explicit execution graphs, reducing latency relative to conventional ROS~2 Python pipelines~\cite{dora_website}. The broader lesson is that runtime architecture matters: dataflow-style execution, reduced interpreter overhead, and explicit dependency graphs are often more compatible with predictable timing than general-purpose task orchestration.

Transport and communication optimizations tackle the same issue from the data path. On unified-memory platforms, bandwidth regulation mechanisms such as MemGuard, BWLOCK, and later dynamic or DSU-level policies protect safety-critical traffic by throttling interferers and reserving DRAM service under contention~\cite{yun2013memguard,7784697,9211564,pradhan2025predictable}. Zero-copy shared-memory transport such as Iceoryx reduces serialization and large-payload copying overhead~\cite{iceoryx_v1_0_0}, while Hazcat and FPGA-assisted DDS reduce host-side transport cost by unifying memory handling or offloading communication across heterogeneous devices~\cite{bell2023hardware,mayoral2024ros}. These methods do not remove timing variability, but they reduce how much of it is injected into the critical path.

When execution spans cloud and edge, communication-aware partitioning becomes necessary. Dynamic split-compute frameworks such as Neurosurgeon and SPINN adapt the partition point to current channel conditions~\cite{kang2017neurosurgeon,teerapittayanon2017distributed}. BottleNet, Elf, and related selective-transmission methods further reduce bandwidth demand through representation compression or sparse token transfer~\cite{eshratifar2019bottlenet,10.1145/3447993.3448628,li2025taskorientedconnectivitynetworkedrobotics}. These approaches improve survivability under communication variability, but they do so by trading model fidelity, freshness, or cross-modal coherence for lower transport load.

Across this family, the conclusion is straightforward: timing reliability in embodied FM deployment is not recovered by faster models alone. It depends on coordinated support across middleware scheduling, kernel preemption, runtime design, bandwidth isolation, and communication-aware execution. The objective is not perfect determinism under all conditions, but an execution stack in which timing disruptions are bounded, localized, and less likely to cascade into unsafe control behavior.
Table~\ref{tab:deployment_solutions_1_4} summarizes the mitigation families that address the long-horizon robustness, safety-assurance, timing, and communication barriers in embodied foundation model deployment.

\begin{table}[t]
\centering
\caption{Mitigation families for the Deployment Gauntlet: representative strategies for improving embodied foundation model deployment at the edge.}
\label{tab:deployment_solutions_1_4}
\scriptsize
\setlength{\tabcolsep}{3pt}
\renewcommand{\arraystretch}{1.12}
\begin{tabularx}{\columnwidth}{|m{1.55cm}|m{2.05cm}|m{2.1cm}|X|}
\hline
\rowcolor[HTML]{800000}
{\color{white}\textbf{Mitigation Family}} &
{\color{white}\textbf{Core Strategy}} &
{\color{white}\textbf{Deployment Benefit}} &
{\color{white}\textbf{Representative Techniques / Evidence}} \\
\hline\hline

Transport-Aware Sensing and Synchronization Relaxation
& Reduce transport overhead, relax rigid synchronization, and preserve usable temporal structure before inference begins.
& Lower tail latency and jitter, reduce copy overhead, and improve multimodal alignment without forcing the stack to wait on the slowest modality.
& Zero-copy middleware transport and loaned buffers~\cite{ishikawaaso2025agnocast,ros2zerocopy_design,ishikawaaso2025agnocast}, executor and synchronization refinements~\cite{kronauer2021latency,abaza2024jitter,teper2024bridging,sobhani2023timing}, variable-sized zero-copy and accelerator-native transport~\cite{ishikawaaso2025agnocast,nitros2025}, pipelined and asynchronous sensing/action generation~\cite{huang2025mmedge,jiang2025asyncvla}, calibration maintenance~\cite{cheng2025calibrefine}, and training-time robustness to misalignment or missing modalities~\cite{song2024timealignmultimodalobjectdetection,wang2024unibevmultimodal3dobject,liu2017learning,wang2025robustadversarialfusiontransformer}. \\
\hline

Hardware-Aligned Model and Compiler Co-Design
& Expose hardware constraints at compile time and redesign operators or models to reduce fallback, fragmentation, and unsupported execution paths.
& Higher accelerator residency, fewer host fallbacks, lower synchronization overhead, and better real-time feasibility on heterogeneous edge hardware.
& SoC-aware compilation and heterogeneous placement~\cite{wali2024hardware,hamdi2024matchmodelawaretvmbasedcompilation,gross2020hardware,lv2025duodecoding,liu2022tinyireemlexecutionenvironment,rognlien2022hardware}, operator reformulation for end-to-end accelerator execution~\cite{wei2025tmanenablingendtoendlowbit,park2025mobilerag}, traffic-reducing quantization and IO-aware attention~\cite{dettmers2022llmint88bitmatrixmultiplication,frantar2023gptqaccurateposttrainingquantization,lin2024awqactivationawareweightquantization,dao2022flashattentionfastmemoryefficientexact,dao2023flashattention,gupta2023flurka}, host-device synchronization reduction~\cite{kwon2023efficientmemorymanagementlarge,zeng2024flightllm}, and hardware-aligned model redesign~\cite{chen2025nanovla,smith2025flexigpt,wen2025tinyvla,leal2023sarartscalingroboticstransformers}. \\
\hline

Memory-Traffic Reduction and Schedulable Memory Use
& Treat shared memory as a schedulable resource and reduce traffic volume so sensing, inference, and control can coexist without excessive jitter.
& More predictable LPDDR service, lower tail latency, fewer stalls, and improved sustained performance under unified-memory contention.
& DRAM isolation and bandwidth policing~\cite{valsan2017addressing,bechtel2024analysis,yun2013memguard,7784697,9211564,fang2024rt,puente2025rosguard,sullivan2024per,zuepke2024mempol}, streaming-stable quantization and pruning~\cite{chen2023lifequant,cai2020zeroq,lin2020hrank,li2023filter,liu2019metapruning,lin2019towards}, outlier suppression and uniform low-bit execution~\cite{ashkboos2024slicegpt,chen2025efficientqat,ashkboos2024quarot,liu2024spinquant}, and KV-cache redesign that trades compute for lower LPDDR pressure~\cite{liu2024kivi,zhang2024kvcache1bit,hooper2025kvquant10millioncontext,shutova2025cachemustadaptivekeyvalue,tomar2025xquant}. \\
\hline

Energy- and Thermal-Aware Execution
& Optimize for sustainable duty cycle and steady-state thermal feasibility rather than short-burst benchmark throughput.
& Lower joules per decision, better endurance, reduced throttling, and more stable long-horizon real-time behavior.
& Endurance-aware deployment evidence~\cite{boroujerdian2018mavbench}, energy-per-inference reduction through distillation and quantization~\cite{sun2020mobilebert,yao2022zeroquant,xiao2023smoothquant,horowitz20141,chen2016eyeriss,koppula2019eden}, duty-cycle reduction for generative control~\cite{chi2025diffusion,wang2024one,wu2025device}, and predictive thermal-aware runtime management~\cite{lee2021thermal,lin2023workload,dowling2023podtas,jacob2025thermml}. \\
\hline

Online Robustness, Adaptation, and Runtime Assurance
& Bound model mismatch through localized correction, constrained adaptation, external safety filters, and runtime supervision rather than unrestricted online relearning.
& Improves long-horizon robustness and reduces unsafe behavior under realistic edge constraints while keeping correction computationally tractable.
& Estimator-level drift compensation through neural bias prediction and lightweight observers~\cite{buchanan2022deep,altawaitan2025learned,kushawaha2025adaptive,george2022closing}, constrained test-time adaptation and streaming-stable updates~\cite{Niu2022EfficientTM,Niu2023TowardsST,Gong2022NOTERC}, sparse and parameter-efficient adaptation~\cite{Zhu2023PockEngineSA,Cai2020TinyTLRM,BenZaken2021BitFitSP}, gradient-free and meta-learned adaptation~\cite{Niu2024TestTimeMA,OConnell2022NeuralFlyER}, control-theoretic safety wrappers and predictive safety filters~\cite{ames2019control,wabersich2021predictive,fisac2018general,cheng2019end,xiao2023safediffuser}, runtime assurance architectures~\cite{desai2019soter,chen2022runtime,tearle2021predictive}, formal verification and uncertainty-aware certificates~\cite{katz2017reluplex,katz2019marabou,ruan2018reachability,tayal2025cp,zhang2025conformal}, and runtime shielding and black-box monitoring~\cite{jansen2020safe,corsi2024verification,zolfagharian2024smarla}. \\
\hline

Real-Time Systems Support and Timing-Aware Communication
& Bound interference across middleware, runtime, kernel, memory, and transport layers so timing disturbances do not cascade across perception, planning, and control.
& Improves timing predictability, reduces transport-induced jitter, and limits the spread of scheduling and communication variability through the execution stack.
& Middleware callback isolation and chain-aware scheduling~\cite{casini2019response,Yang2023FoundationMF,choi2021picas}, kernel-level real-time support~\cite{gutierrez2018real}, compiled dataflow runtimes and zero-copy execution graphs~\cite{dora_website}, DRAM bandwidth regulation and arbitration control~\cite{yun2013memguard,7784697,9211564,pradhan2025predictable}, zero-copy and heterogeneous I/O transport optimization~\cite{iceoryx_v1_0_0,bell2023hardware,mayoral2024ros}, and communication-aware cloud--edge partitioning with compression or selective transmission~\cite{kang2017neurosurgeon,teerapittayanon2017distributed,eshratifar2019bottlenet,10.1145/3447993.3448628,li2025taskorientedconnectivitynetworkedrobotics}. \\
\hline

\end{tabularx}
\end{table}
\FloatBarrier
\vspace{-.4cm}

\section{Future Directions: Beyond the Deployment Gauntlet}
\label{sec:future_roadmap}

\begin{figure}[t!]  
   \centering
   \includegraphics[width=1\textwidth]{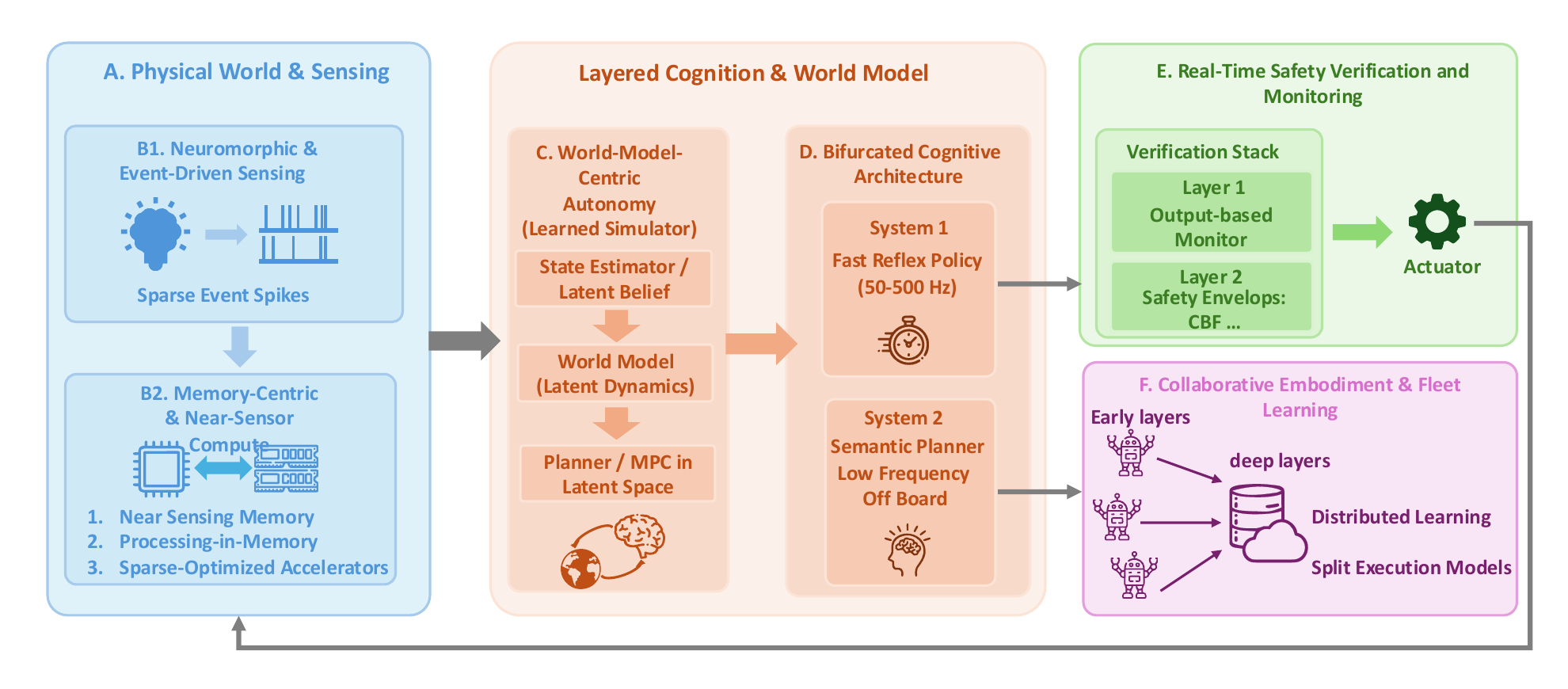} 
   \caption{Architecture trends for future Embodied Foundation Models: Data volume is minimized via Neuromorphic and event sensing (A), and near sensor compute mechanism (B). A central predictive world model stabilizes perception (C) while cognition is separated into fast reflexes and slow reasoning (D). The rigorous safety policy(E) enforces constraints before actuation. And the Fleet learning (F) extends the feasible complexity.}
  \label{figure-futurework}
\end{figure}

The Deployment Gauntlet suggests that incremental gains in compression, quantization, or model scaling often improve \emph{average} performance more than \emph{worst-case} deployment behavior under SWaP-constrained, closed-loop execution. Many of the most consequential failures-including memory contention, missed deadlines, thermal excursions, and safety-envelope violations-arise from cloud-inference assumptions such as best-effort scheduling, batch-oriented memory traffic, and weak timing contracts. Figure~\ref{figure-futurework} summarizes a set of representative architectural responses to these constraints. 

\vspace{0.25em}
\noindent\textbf{Representative architectural directions:}
\begin{itemize}\setlength{\itemsep}{0.2em}
\item \textbf{Event-driven sensing:} Replace dense frame pipelines with event-structured inputs to reduce redundant compute and bandwidth, improve temporal alignment, and tighten end-to-end latency budgets~\cite{ali2024energy, zheng2023deep}.
\item \textbf{Memory-centric and near-sensor compute:} Reorganize dataflow to reduce tensor movement through near-sensor preprocessing and PIM-style primitives, mitigating the unified-memory contention that often dominates edge latency and energy~\cite{Kanjo2026neuromorphic}~\cite{aalishah2025edgenavmamba}.
\item \textbf{World-model-centric autonomy:} Use predictive state as a synchronization substrate under delay, occlusion, and sensor disagreement, enabling temporally grounded planning rather than purely reactive control~\cite{wu2023daydreamer, feng2025survey}.
\item \textbf{Bifurcated cognition (System~1 / System~2):} Separate high-rate control authority from lower-rate semantic reasoning so that deliberative stalls do not propagate directly into actuation hazards~\cite{Ahn2022DoAI, Firoozi2023FoundationMI}.
\item \textbf{Real-time safety enforcement:} Combine lightweight monitoring with bounded-overhead runtime constraint enforcement, such as CBF-style reflex layers, that remain effective under missed deadlines and resource pressure~\cite{tolle2023runtime, Firoozi2023FoundationMI}.
\item \textbf{Collaborative embodiment and fleet learning:} Treat robustness partly as a fleet-level property by amortizing rare events and distribution shift across deployments through federated, split, or distilled update mechanisms under limited connectivity~\cite{wu2023efficient}.
\end{itemize}

\subsection{Evaluation Trends: From Accuracy to Operational Validity}

These architectural directions imply a corresponding shift in evaluation. Accuracy and average latency remain necessary, but they are weak predictors of real deployment outcomes when the execution stack is constrained by timing, memory, and thermal pressure. Future benchmarks should therefore report at least three additional properties: (i) \textbf{temporal coherence}, including state freshness and synchronization under load; (ii) \textbf{bounded failure and recoverability}, including graceful degradation within safety envelopes; and (iii) \textbf{resource stability}, including sensitivity to bandwidth pressure, I/O contention, thermal throttling, and scheduling jitter. Stress-test protocols that inject jitter, bus pressure, packet loss, and sensor dropout would make evaluation more operationally meaningful by measuring whether systems remain coherent, bounded, and safe under realistic constraints.

\section{Conclusion}

Deploying foundation models on edge embodied platforms is a systems problem, not simply a problem of compression or accelerator efficiency. Reliable embodied performance depends on the interaction of sensing, computation, memory, timing, power, and safety under closed-loop execution. We organized these recurring constraints through the Deployment Gauntlet, a systems taxonomy that explains why gains in average-case throughput do not necessarily translate into dependable deployment. Across the literature, quantization, pruning, and related optimizations improve deployability, but they do not, on current evidence, resolve the worst-case failures associated with memory contention, timing variability, thermal instability, and limited safety assurance.

A central implication of this survey is that embodied deployment increasingly favors architectural decomposition over monolithic end-to-end execution. Separating high-rate control from lower-rate semantic reasoning is one promising response to the tension between control-loop determinism and deliberative inference. More broadly, reducing data movement is often as important as reducing arithmetic cost. This makes transport-aware sensing, memory-centric execution, and near-sensor or in-memory processing especially relevant for future embodied systems operating under strict SWaP constraints.

The broader lesson is that robust embodied intelligence will depend less on transferring cloud-oriented model assumptions to the edge and more on explicit co-design across sensing, memory hierarchy, runtime scheduling, communication, and model architecture. The semantic capabilities of foundation models remain valuable, but realizing them in physical systems requires stacks designed around edge constraints rather than adapted to them only after training. In that sense, the future of embodied foundation models lies not simply in smaller models on smaller devices, but in systems whose learning objectives, execution substrate, and safety requirements are aligned from the outset.

\bibliographystyle{ACM-Reference-Format}
\bibliography{sample-base}
\end{document}